\documentclass[12pt]{spieman}  
\usepackage{amsmath,amsfonts,amssymb}
\usepackage{graphicx}
\usepackage{setspace}
\usepackage{tocloft}

\usepackage{color}
\usepackage{caption}
\usepackage{wrapfig}
\usepackage{multirow, makecell}
\usepackage{booktabs}

\newcommand{\eg}{\textit{e.g.}}
\newcommand{\etal}{\textit{et al.~}}

\title{Self-Supervised Intensity-Event Stereo Matching}

\author[a,$\dagger$]{Jinjin Gu}
\author[b,$\dagger$]{Jinan Zhou}
\author[c,$\dagger$]{Ringo Sai Wo Chu}
\author[c]{Yan Chen}
\author[c]{Jiawei Zhang}
\author[c]{Xuanye Cheng}
\author[c]{Song Zhang}
\author[c,d]{Jimmy S. Ren}

\affil[a]{School of Electrical and Information Engineering,
The University of Sydney}
\affil[b]{Carnegie Mellon University}
\affil[c]{SenseTime Research and Tetras.AI}
\affil[d]{Centre for Perceptual and Interactive Intelligence}

\cftpagenumbersoff{figure}
\cftpagenumbersoff{table} 
\begin{document} 
\maketitle

\begin{abstract}
Event cameras are novel bio-inspired vision sensors that output pixel-level intensity changes in microsecond accuracy with a high dynamic range and low power consumption.
Despite these advantages, event cameras cannot be directly applied to computational imaging tasks due to the inability to obtain high-quality intensity and events simultaneously.
This paper aims to connect a standalone event camera and a modern intensity camera so that the applications can take advantage of both two sensors.
We establish this connection through a multi-modal stereo matching task.
We first convert events to a reconstructed image and extend the existing stereo networks to this multi-modality condition.
We propose a self-supervised method to train the multi-modal stereo network without using ground truth disparity data.
The structure loss calculated on image gradients is used to enable self-supervised learning on such multi-modal data.
Exploiting the internal stereo constraint between views with different modalities, we introduce general stereo loss functions, including disparity cross-consistency loss and internal disparity loss, leading to improved performance and robustness compared to existing approaches.
The experiments demonstrate the effectiveness of the proposed method, especially the proposed general stereo loss functions, on both synthetic and real datasets.
At last, we shed light on employing the aligned events and intensity images in downstream tasks, \textit{e.g.}, video interpolation application.
\end{abstract}


{\noindent \footnotesize\textbf{*} Jimmy S. Ren,  \linkable{jimmy.sj.ren@gmail.com} }

{\noindent \footnotesize\textbf{$\dagger$} Work was done when they were interns at sensetime research.}

\begin{spacing}{1}   

\section{Introduction}
\label{sect:intro}  
Event cameras measure changes in brightness at each pixel independently instead of reporting pixel activations.
Event cameras have attracted increasing attention for their high temporal resolution, high dynamic range and low power consumption features and have been applied to various computer vision tasks \cite{rebecq2016evo,jiang2020learning,zhu2018ev,vidal2018ultimate,tulyakov2021time}.
However, existing event cameras are either unable to obtain high-quality image pixel intensities (DVS \cite{patrick2008128x} sensors only output events) or suffer low spatial resolution and lack of color information (dynamic and active-pixel vision sensors \cite{brandli2014240}).
These limitations make it difficult for event cameras to assist computational imaging tasks, as we cannot obtain high-resolution intensity images and events simultaneously.

\begin{figure}[t]
    \centering
    \includegraphics[width=\linewidth]{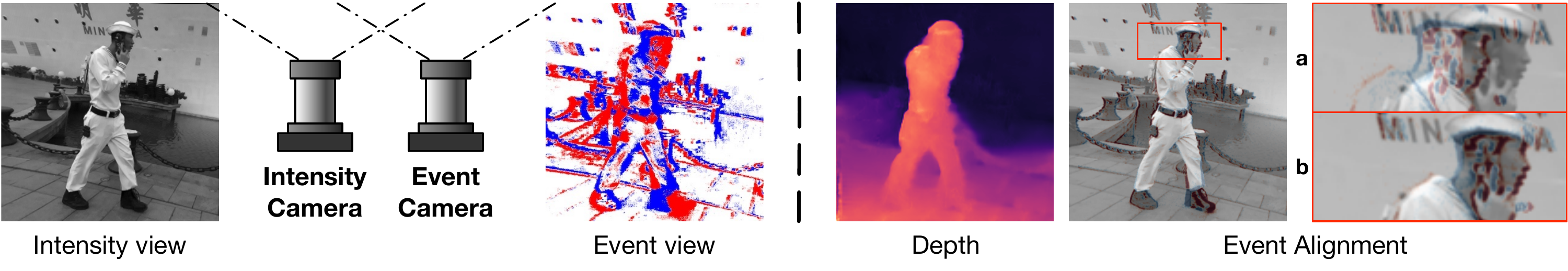}
    \caption{The proposed intensity-event stereo setting, in which we use an event camera and an intensity camera. With the proposed self-supervised stereo matching model, we can not only obtain the disparity used to calculate the depth map, but also build a connection between these two sensors. (a) The signals are from these two displaced sensors and are unaligned. (b) We align these two signals by the proposed method.}
    \label{fig:cover}
\end{figure}

In this paper, we aim to connect a standalone event camera and a separate modern intensity camera so that applications could exploit the advantages of both of the sensors (see \figurename~\ref{fig:cover}).
Such application scenarios are not uncommon for most consumer-level imaging devices simply because acquiring colorful visual contents with high resolution, high speed, and low power consumption is without the scope of any individual image sensors.
We establish a connection between these two sensors through a computational stereo matching model and estimate their disparity.
This disparity describes the relationship between these two sensors and allows the sensors to be combined to complete the task that one sensor cannot achieve, \eg, obtaining both high-resolution images and events simultaneously for downstream tasks.

However, studying this problem is NOT a naive extension of the existing stereo matching methods on a new sensor setting, owing to the following technical barriers.
Firstly, the current stereo networks are not optimal for multi-modal problems. They assume that left and right view images have the same modality and use the weights shared feature extraction model for these images.
Secondly, it is challenging to obtain multi-modal data for training.
Especially, the acquisition of ground truth disparity for each different setting is expensive.
To practically apply a multi-modal stereo framework, we need a robust training strategy and get rid of the shackles of data annotation.
On this front, self-supervised learning provides a promising perspective to use the inherent constraints and characteristics of the data to learn the desired stereo matching without ground truth disparity indirectly.
However, the prerequisite for the success of the existing self-supervised stereo matching framework is to establish the photometric consistency relationship between the projected images from two views.
This brings up the third problem.
The left and right view signals have different physical meanings and data structures in our setting.
This causes the failure of the self-supervised learning framework as the previous photometric constraint does not hold.

In this work, we propose a self-supervised method for learning the multi-modal stereo matching without any ground truth disparity.
To facilitate the existing outstanding image stereo models on the proposed intensity-event setting, we first convert the event stream to roughly reconstructed images through the off-the-shelf models \cite{rebecq2019events,scheerlinck2020fast}.
The roughly reconstructed images are still in a different modality from the images of the other view as the color and detail information can not be well restored.
We improve the existing stereo networks and make images with different modalities to use modality-specific feature extraction sub-modules.
In the proposed self-supervised method, we introduce a gradient structure consistency loss for the geometry constraints between the intensity and the reconstructed images after projection, which mainly leverages the edge information provided by events.
Last but not least, only using the structure consistency may result in poor quality disparity maps as the supervision is sparse and vague.
To overcome this issue, we propose a novel loss based on the cross-consistency between the disparities calculated across different views using different modality images.
We also constraint our training according to the fact that the disparity of the same view should be zero.
The proposed loss functions lead to improved stereo matching performance and robustness.

The calculated disparity maps can be used in many computational photography tasks, with depth estimation first.
Projecting events to intensity camera view also allows many applications that could not be realized in the past due to hardware limitations.
We can now obtain high-resolution events and intensity images simultaneously.
At last, we experimentally demonstrate the potential of the proposed framework using the warped event to facilitate event-based video frame interpolation task.

\section{Related Work}
\label{sec:related}

\subsection{Event Cameras.}
Event camera is a kind of sensor that records signals when the scene exhibits illumination changes \cite{patrick2008128x,brandli2014240}.
An event camera reports signals (events) asynchronously when the log intensity change exceeds a preset threshold $\tau$.
We have witnessed the rise of event cameras due to their distinctive advantages over conventional active pixel cameras, \eg, higher frame rate, higher dynamic range and lower power consumption.
These properties attracted the use of event cameras in many computer vision tasks, \eg, tracking  \cite{rebecq2016evo,gehrig2018asynchronous}, deblurring \cite{jiang2020learning,lin2020learning}, optical flow estimation \cite{zhu2019unsupervised,zhu2018ev}, SLAM \cite{kim2016real,kueng2016low,vidal2018ultimate}, video frame interpolation \cite{lin2020learning,tulyakov2021time}.
However, the unique data structure of event cameras renders the existing computer vision tools and algorithms unusable, which places a major obstacle against the application of event cameras.
Many works have been focusing on bridging events and conventional cameras by reconstructing intensity frames from events \cite{barua-event2img,bardow,mostafavi_cgan_event_image_translation,mostafavi2020e2sri,Wang_2020_CVPR_eventSR,stoffregen2020reducing}, thus allowing modern vision algorithm to take place.
Rebecq \etal \cite{rebecq2019events} propose E2VID, a recurrent network to reconstruct videos from a stream of events and train it on a large amount of simulated event data. 
Scheerlinck \etal \cite{scheerlinck2020fast} propose FireNet, which simplifies the neural architecture in \cite{rebecq2019events} with a smaller number of parameters while maintaining similar quantitative results.
Despite that many works attempts to reconstruct the intensity image from the event, none of these methods can recover the intensity and color information well.
Therefore, the absence of color information in the reconstructed image degrades the performance for downstream tasks.
In our application, the color mismatch makes the existing self-supervised stereo matching algorithm based on photometric consistency invalid.

\begin{figure}[t]
    \centering
    \includegraphics[width=\linewidth]{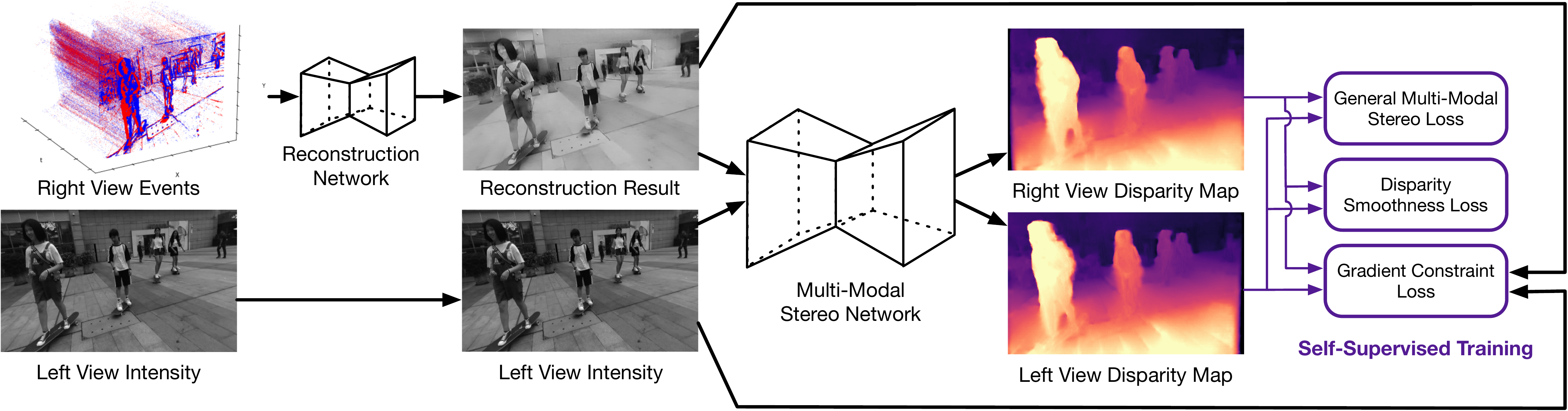}
    \caption{This figure shows the overall framework of the proposed method. Firstly, we obtain rough reconstruction results using the existing methods from the right view events. The multi-modal stereo network predicted a disparity map based on the right reconstructed results and left intensity images. The whole system uses three kinds of loss functions for self-supervised training.}
    \label{fig:framework}
\end{figure}

\subsection{Stereo Matching.}
Stereo matching is the process of linking the pixels in different views that correspond to the same point of the scene.
It follows a long line of research works.
Early works involve searching and matching correspondence pixels on the epipolar line \cite{Yoon_stereo_early,hosni_stereo_early}
Recently, deep learning based methods have dominated the field of stereo matching due to their superior performance and usability.
Zbontar and LeCun \cite{Zbontar_2015_CVPR} are among the first to use a convolutional network for computing stereo matching cost in image pairs.
After that, a lot of work was proposed to improve the performance, \eg, inner product layer \cite{luo2016efficient}, encoder-decoder architecture \cite{mayer2016large}, 3D convolution cost-volume module \cite{Kendall_2017_ICCV}, spatial pyramid pooling and 3D hourglass convolution \cite{Chang_2018_CVPR}, guided attention cost-column \cite{Zhang_2019_GANet}, PatchMatch module for sparse cost volume representation \cite{duggal2019deeppruner}, intra-scaling cost aggregation \cite{xu2020aanet}.
With the development of various sensors, multi-model and cross-spectral stereo matching has become an emerging topic \cite{chiu2011improving,zhi2018deep,shen2014multi,jeon2016stereo,kim2015dasc,kim2016deep}.
But none of them is suitable for calculating the correspondence between intensity images and events or event reconstruction images.
Concurrent with our work, Mostafavi \etal \cite{Mostafavi_2021_ICCV} investigate stereo matching with event-intensity cameras on both views and propose an event-intensity network that refines image details using events.
Our work is essentially different in purpose and method; we only use one intensity and one event camera and train our model self-supervised.

\subsection{Self-Supervised Learning}
Learning-based stereo methods are data-hungry. They often require a lot of ground truth data for training.
Over the past few years, self-supervised models have been developed to learn stereo matching without ground truth annotations.
They are usually built on the principles of disparity smoothness prior and re-projection photometric consistency constraints.
Garg \etal \cite{garg2016unsupervised} tackle monocular depth estimation by minimizing the loss between the source image and backwards-warping from the subsidiary stereo image.
Similarly, Godard \etal \cite{godard2017unsupervised} include a left-right consistency to enforce disparity prediction.
They further propose a new minimum re-projection loss and auto-masking loss to improve the performance \cite{godard2019digging}. 
Zhou \etal \cite{zhou2017unsupervised} adopt left-right check to guide the training and pick suitable matching as training data.
Zhi \etal \cite{zhi2018deep} propose a self-supervised learning framework for cross-spectral stereo matching.
They introduce a material-aware loss function to handle regions with unreliable matching.
However, their method involves the translation between intensity and near-infrared images, thus unsuitable for our setting.

\section{Method}
\label{sec:method}
In this section, we describe our self-supervised intensity-event stereo matching framework.
We first introduce the problem formulation and overall framework design in Sec \ref{sec:method:framework}.
We then describe the modified stereo network for multi-modal problem in Sec \ref{sec:method:stereo}.
The loss functions are introduced in Sec \ref{sec:method:ssl}, featuring a gradient structure consistency loss and general losses for multi-modal stereo matching.


\begin{figure}[t]
    \centering
    \includegraphics[width=\linewidth]{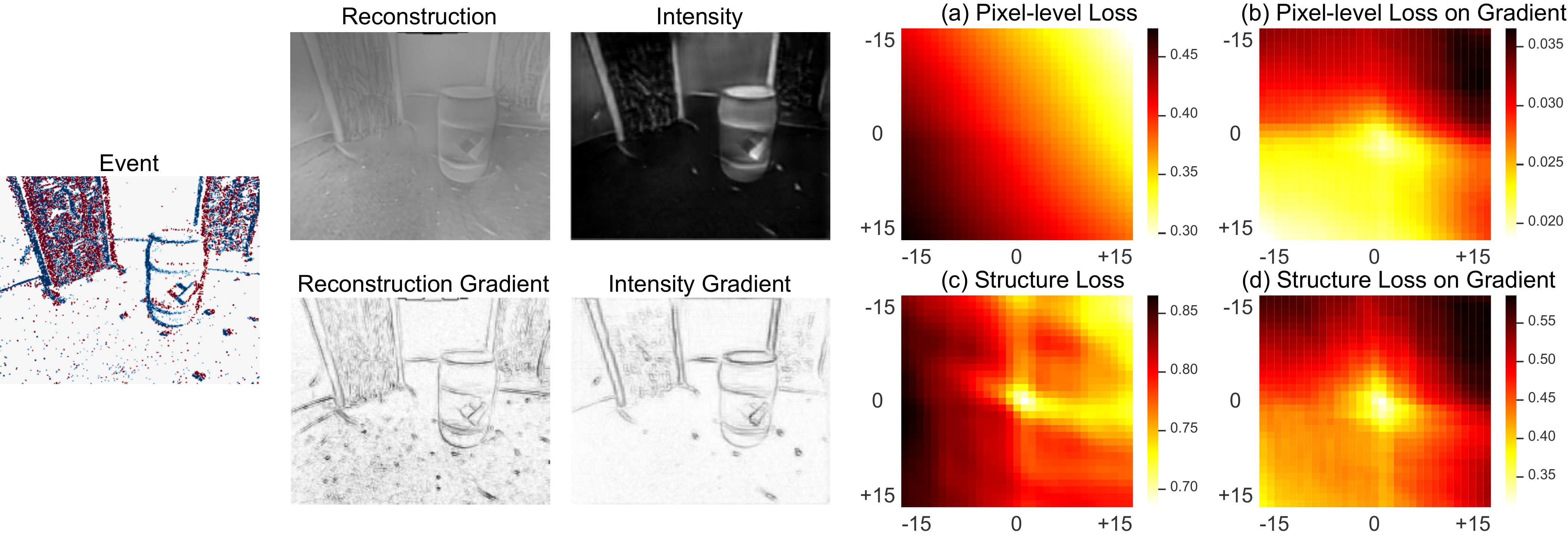}
    \caption{The visualization of different loss functions. On the left of the figure is the event, the reconstructed image using E2VID \cite{rebecq2019events}, the intensity image and their gradient visualization. The heat maps show the loss values with different displaced pixel numbers. In the centre of the heat maps, there is no displacement between the two images being compared, and the loss value should be the smallest.}
    \label{fig:loss}
\end{figure}

\subsection{Overall Framework}
\label{sec:method:framework}
The multi-modal intensity-event stereo matching problem is first formulated as follow.
As shown in \figurename \ref{fig:cover}, an intensity camera and an auxiliary horizontally displaced event camera are used in our setting.
In this work, we assume that the camera on the left is the image camera, and the one on the right is the event camera.
Let $I^l$ be the left view intensity image and $\{E_m^r\}_{m\in \mathbb{N}}$ be the event stream obtained by the right event camera within a short amount of time before the intensity image is captured.
The underlying problem can be considered a data association problem, that is, to find correspondences between the points in the left image and right events.
The correspondences are formed as the final disparity map, which is also the output of the stereo matching problem.

However, the right signal (event) is not in the same modality as the left signal (intensity images), failing the existing methods with the given problem setting.
A reconstruction network is employed at the first to convert the event stream $\{E_m^r\}_{m\in \mathbb{N}}$ to a roughly reconstructed image $I^r$.
In our work, we use two popular event reconstruction models, E2VID \cite{rebecq2019events,rebecq2019high} and FireNet \cite{scheerlinck2020fast}.
Note that these models are replaceable.
Given these two images, we can adapt the existing image stereo models to predict disparity between images, which is also the disparity between the left image and the right events.
However, the event camera does not record the value of the pixels and only record the pixels changing.
Thus the rough reconstruction $I^r$ only contain usable edge information but are unreliable in color and detail and still in different modalities with $I^l$ (see the reconstruction result in \figurename~\ref{fig:loss}).


\subsection{Multi-Modal Stereo Network}
\label{sec:method:stereo}
As stated above, we need a multi-modal stereo network to handle inputs with different modalities (either the right view is event voxel or reconstructed images).
Although we can apply the previous convolutional stereo matching networks theoretically, the difference in modalities still poses challenges.
Most stereo networks are composed of feature extraction, correlation and aggregation sub-models, and the feature extraction model usually share weights for both two views.
This weight sharing strategy is effective originally but poses limitations for images with different modalities.
We make the minor changes to these networks to make images with different modalities using modality-specific feature extraction sub-modules.
This design has two advantages.
Firstly, the feature extraction models dedicated to different modalities avoid confusion between different images.
Secondly, the new design allows us to swap the modalities of the left and right views and predict disparity of the other view, as long as we swap them together with the feature extraction branches.
The second property is essential for the cross-consistency loss, which we will describe in Sec \ref{sec:method:ssl}.
Note that this structure also allows the event voxel to be used directly as input.

\begin{figure}[t]
    \centering
    \includegraphics[width=\linewidth]{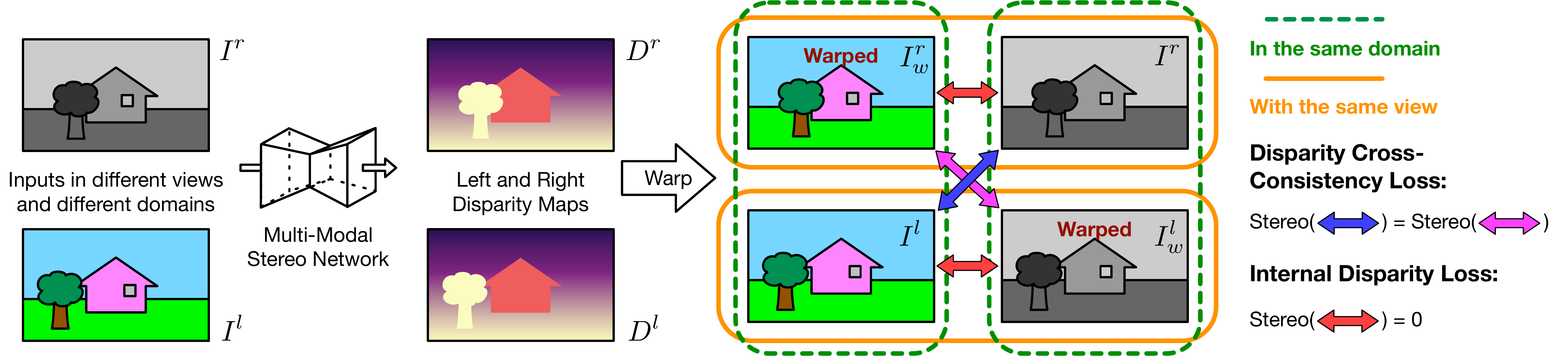}
    \caption{Using projection method and the multi-modal stereo model, we can obtain all four combinations of two views and two modalities. The proposed general multi-modal stereo losses are derived from the geometry constraints between these images, including disparity cross-consistency loss and internal disparity loss.}
    \label{fig:general_loss}
\end{figure}

\begin{figure}[t]
    \centering
    \includegraphics[width=\linewidth]{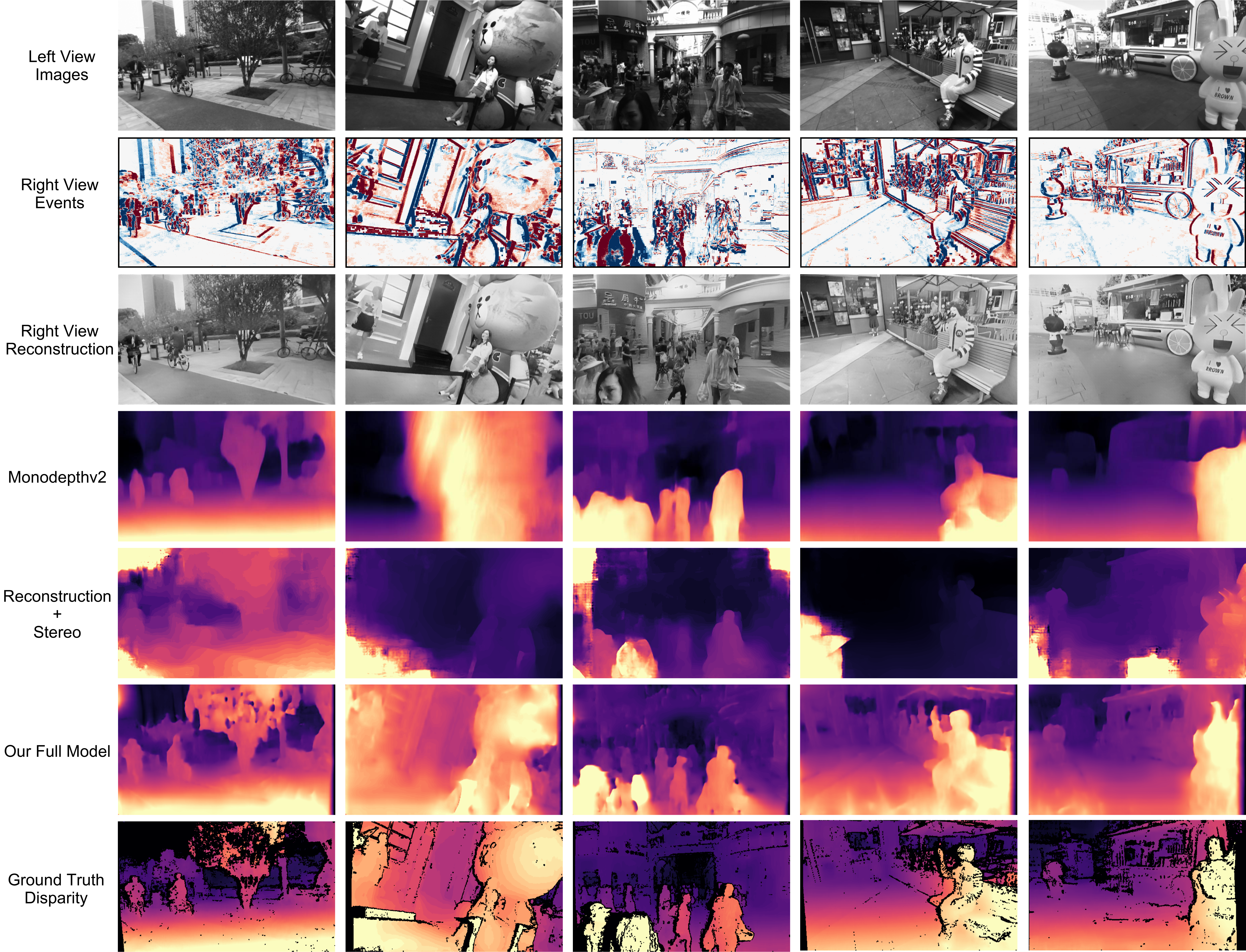}
    \caption{Results of different methods on synthetic dataset. Monocular model does not work well as it cannot be fine-tuned under this setting. Stereo matching between intensity and reconstruction fails because the color discrepancy prevents the network to relate corresponding pixels. After self-supervised training, our predictions are on par with the ground truth. The reconstruction network is FireNet and the stereo network is the modified AANet.}
    \label{fig:main_synthetic}
\end{figure}

\begin{figure}[t]
    \centering
    \includegraphics[width=\linewidth]{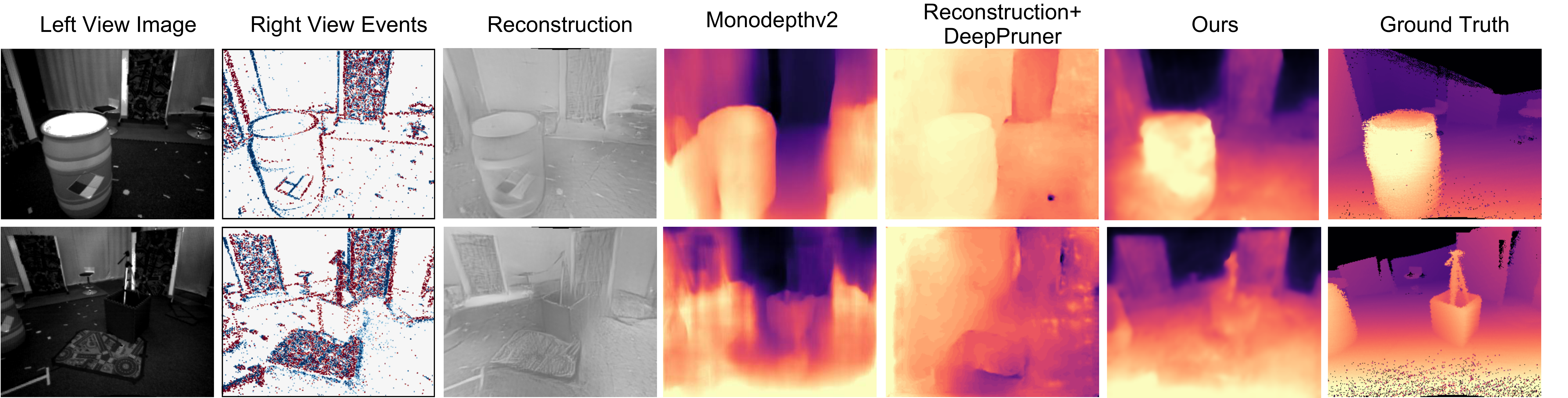}
    \caption{Comparison of different methods on the MVSEC dataset \cite{zhu2018multivehicle}. Notice that the direct event reconstruction quality on real data is far inferior to that of synthetics data, yet our framework can still achieve decent results, which shows its robustness. The reconstruction network is E2VID, and the stereo network is the modified AANet.}
    \label{fig:comp_mvsec}
\end{figure}

\subsection{Loss Function}
\label{sec:method:ssl}
To achieve the goal of self-supervised learning, we define our loss function as the combination of the following four parts:
\begin{equation}
    \mathcal{L}=\lambda_{gd}\mathcal{L}_{gd}+\lambda_{sm}\mathcal{L}_{sm}+\lambda_{cc}\mathcal{L}_{cc}+\lambda_{itn}\mathcal{L}_{itn},
\end{equation}
where the $\lambda$s denote the hyper-parameters that control the loss weights.
Next, we describe each component of our loss, respectively.

\subsubsection{Gradient Structure Consistency Loss.}
One core guarantee for the success of self-supervised stereo matching is that the output disparities indicate the epipolar geometry relationship between the left and right views.
Usually, we can achieve this goal by comparing the projected left image and right images with image similarity metrics, \eg, pixel-wise loss and perceptual loss \cite{zhang2018unreasonable,jinjin2020pipal}.
However, the quality of the reconstruction result is usually poor, which fails the previous loss functions.
\figurename~\ref{fig:loss} shows an example of reconstruction.
As one can observe, the reconstruction network fails to recover any color information. However, its gradient reserves the structure information of the scene.
We propose to use image structure loss \cite{wang2004image} calculated on image gradient to constraint the stereo training.
Let $G^l=\nabla_{xy} I^l$ be the gradient of $I^l$ and $G^r=\rho\nabla_{xy} I^r$ be the adjusted gradient of $I^r$ with scaling factor $\rho$.
The used gradient structure consistency loss is formulated as
\begin{equation}
    \mathcal{L}_{gd}=1-\frac{2\mu_{G^l}\mu_{G^r}+c_1}{\mu_{G^l}^2+\mu_{G^r}^2+c_1}\times\frac{2\sigma_{G^lG^r}+c_2}{\sigma_{G^l}^2+\sigma_{G^r}^2+c_2},
    \label{eq:ssim}
\end{equation}
where $c_1$, $c_2$ are constants and $\mu_{G^l}$, $\mu_{G^r}$, $\sigma_{G^l}$, $\sigma_{G^r}$, $\sigma_{G^lG^r}$ represent means, standard deviations and cross-covariance of the gradient pair.
In practice, Eq \eqref{eq:ssim} is calculated on the local patch pairs and then summed up as the final loss.

We compare different losses in \figurename~\ref{fig:loss}.
We set the reconstructed image unmove and shift the intensity images so that they are unaligned to simulate the situation after projection during self-supervised training.
We then visualize the distribution of the loss values.
We examine commonly used pixel-wise $l_1$-norm loss \cite{godard2017unsupervised}, pixel-wise $l_1$-norm loss on the image gradient, structure loss \cite{godard2017unsupervised} on the image, and the proposed structure loss on the image gradient.
As one can observe, the pixel-wise losses cannot indicate the optimal point.
The structure loss calculated on images can indicate the optimal point but has an unsmooth loss landscape.
Only the proposed loss has a relatively smooth loss landscape while successfully indicating the optimal point.

Directly calculating photometric consistency between warped images may introduce blurring to the predicted disparity map because there are occlusion areas between left and right view scenes, where warping cannot fill.
We introduce the occlusion mask $M$ to mask out these occlusion pixels.
We first perform left-right consistency check by projecting the right disparity using the left disparity map and calculate their coherence.
The inconsistent region, which is likely to be the occlusion region, is marked as the occlusion mask $M$, which can be formulated as
\begin{equation}
    M=\begin{cases}
    0,\quad\|D^l - P(D^r; D^l)\|_1<t\\
	1,\quad\|D^l - P(D^r; D^l)\|_1\geq t
  \end{cases},
\end{equation}
where $P(D^r; D^l)$ represents projecting the right disparity using the left disparity and $t$ is the threshold parameter.

\subsubsection{Disparity Smoothness Loss.}
After obtaining the disparity maps for both views $D^l$ and $D^r$, we follow the previous methods for estimating dense flow or disparity \cite{godard2017unsupervised} and employ an edge-ware smoothness loss.
This loss is symmetrical for both left and right views. Thus we omit the superscript.
We encourage disparities to be locally smooth with a penalty on the disparity gradients $\nabla_x D$ and $\nabla_y D$.
As depth discontinuities often occur at image edges, we weight this cost with an edge-aware term using the image gradients $\nabla_x I$ and $\nabla_y I$, which is formulated as
\begin{equation}
    \mathcal{L}_{\mathrm{sm}}=\frac{1}{N}\sum_{i,j}|\nabla_x D_{ij}|e^{-|\nabla_x I_{ij}|}+|\nabla_y D_{ij}|e^{-|\nabla_y I_{ij}|},
\end{equation}
where $D$ denoted the disparity map corresponded with $I$, the subscripts $i$ and $j$ indicates pixel coordinates, $N$ is the total number of pixels.

\begin{table}[t]
    \centering
    \resizebox{0.8\linewidth}{!}{
    \begin{tabular}{p{7cm}|p{1.2cm}<{\centering}p{1.2cm}<{\centering}p{1.2cm}<{\centering}p{1.2cm}<{\centering}}
    \toprule
        \multirow{2}{*}{Model} & \multirow{2}{*}{EPE $\downarrow$} & \multicolumn{3}{c}{Bad Pixels $\downarrow$} \\
         & & $\delta>1$ & $\delta>3$ & $\delta>5$ \\
    \midrule
        Monodepth2 & 8.849 & 0.953 & 0.781 & 0.648\\
        \midrule
        DeepPruner (upper bound) & 0.712 & 0.123 & 0.027 & 0.015 \\
        \midrule
        FireNet+AANet (baseline) & 4.811 & 0.649  & 0.419  & 0.336 \\
        E2VID+AANet (baseline) & 5.154 & 0.673  & 0.440  & 0.379\\
        FireNet+DeepPruner (baseline) & 10.29 & 0.417 & 0.226 & 0.181 \\
        E2VID+DeepPruner (baseline) & 6.386 & 0.381 & 0.184 & 0.140 \\
        \midrule
        FireNet+AANet* ($\mathcal{L}_{gd}$ and $\mathcal{L}_{sm}$)  & 1.591  & 0.366 & 0.139 & 0.088 \\
        E2VID+AANet* ($\mathcal{L}_{gd}$ and $\mathcal{L}_{sm}$) & 1.496  & 0.351 & 0.123 & 0.075\\
        FireNet+DeepPruner* ($\mathcal{L}_{gd}$ and $\mathcal{L}_{sm}$) & 1.336 & 0.355 & 0.123 & 0.068 \\
        E2VID+DeepPruner* ($\mathcal{L}_{gd}$ and $\mathcal{L}_{sm}$) & 1.321 & 0.355 & 0.116 & 0.068 \\
        \midrule
        FireNet+AANet (all losses)  & 1.988 & 0.409 & 0.189 & 0.134 \\
        E2VID+AANet (all losses) & 1.775 & 0.378 & 0.166 & 0.117 \\
        FireNet+DeepPruner (all losses) & 1.626 & 0.377 & 0.147 & 0.097\\
        E2VID+DeepPruner (all losses) & 1.57 & 0.368 & 0.143 & 0.094 \\
        \midrule
        FireNet+AANet* (all losses)  & 1.201 & 0.306 & 0.110 & 0.065 \\
        E2VID+AANet* (all losses) & 1.101 & 0.287 & 0.094 & 0.057 \\
        FireNet+DeepPruner* (all losses) & 0.971 & 0.317 & 0.087 & 0.049\\
        E2VID+DeepPruner* (all losses) & 0.913 & 0.289 & 0.074 & 0.042 \\
    \bottomrule
    \end{tabular}
    }
    \caption{Quantitative comparison of different approaches on stereo matching using our synthetic stereo event dataset \cite{zhou2019davanet}. $\uparrow$ means the higher the better while $\downarrow$ means the lower the better. ``*'' indicates using the modified stereo network.}
    \label{tab:comparison}
\end{table}

\subsubsection{General Multi-Modal Stereo Losses.}
Although gradient structure consistency loss can guide stereo training, the provided supervision is sparse and not that specific as pixel-level losses.
We can not obtain accurate disparity only with the above losses.
Exploiting the internal stereo relationship between different views and different modalities, we propose general multi-modal stereo losses.
An simple illustration of the proposed losses is shown in \figurename \ref{fig:general_loss}.
With the $I^l$ and $I^r$ at one hand, we calculate the disparities $D^l$ and $D^r$ that correspond to $I^l$ and $I^r$, respectively.
By projecting $I^l$ and $I^r$ according to $D^l$ and $D^r$, we obtain $I^r_\mathrm{w}$ and $I^l_\mathrm{w}$, which represent different views and are in different modalities:
$I^l_\mathrm{w}$ is with the same modality with $I^r$ but with the same view with $I^l$; and $I^r_\mathrm{w}$ is with the same modality with $I^l$ but with the same view with $I^r$.
Using the same multi-modal stereo network, we can obtain $D^l_\mathrm{w}$ and $D^r_\mathrm{w}$ -- the disparities calculated on two projected images.
The proposed loss functions are built based on two facts.
According to the fact that the disparity between $I^l_\mathrm{w}$ and $I^r_\mathrm{w}$ should be the same disparity between $I^l$ and $I^r$, we build the disparity cross-consistency loss to make them as close as possible:
\begin{equation}
    \mathcal{L}_{cc}=\frac{1}{N}\sum_{i,j}\Big||D^l|-|D^r_\mathrm{w}|\Big|+\Big||D^r|-|D^l_\mathrm{w}|\Big|,
\end{equation}
where we take the absolute value for disparities as the projection directions may be opposite, and we only need their shapes.
According to another fact that there should be no disparities within the same view, we build the internal disparity loss:
\begin{equation}
    \mathcal{L}_{\mathrm{itn}}=\frac{1}{N}\sum_{i,j}|D^r_{\mathrm{itn}}|+|D^l_{\mathrm{itn}}|,
\end{equation}
where $D^r_{\mathrm{itn}}$ is the calculated between $I^r$ and $I^r_\mathrm{w}$ and $D^l_{\mathrm{itn}}$ is the calculated between $I^l$ and $I^l_\mathrm{w}$.

\begin{table}
	\centering
	\footnotesize
    \begin{tabular}{l|cccc}
    \toprule
        \multirow{2}{*}{Model} & \multirow{2}{*}{EPE $\downarrow$} & \multicolumn{3}{c}{Bad Pixels $\downarrow$} \\
         & & $\delta>1$ & $\delta>3$ & $\delta>5$ \\
    \midrule
        Monodepth2 & 10.235 & 0.914 & 0.844 & 0.768\\
        E2VID+AANet (baseline) & 11.332 & 0.954  & 0.864  & 0.776\\
        E2VID+AANet (all losses) & 5.830 & 0.736 & 0.660 & 0.434 \\
        E2VID+DeepPruner (all losses) & 4.979 & 0.673 & 0.581 & 0.384 \\
        E2VID+AANet* (all losses) & 2.734 & 0.653 & 0.330 & 0.197 \\
        E2VID+DeepPruner* (all losses) & 2.397 & 0.601 & 0.268 & 0.164 \\
    \bottomrule
    \end{tabular}
	\caption{Quantitative comparison of different approaches on stereo matching using real-world dataset MVSEC \cite{zhu2018multivehicle}. $\uparrow$ means the higher the better while $\downarrow$ means the lower the better. - means the method completely fails.}
	\label{tab:comparison_real}
\end{table}

\section{Experiments}
\label{sec:exp}

\subsection{Implementation Details}
\label{sec:exp:imp}
In this section, we experimentally evaluate the stereo matching performance of the proposed method.
We use both synthetic data and real data in our experiments.
For the experiments based on synthetic data, we employ the Stereo Blur Dataset proposed by Zhou \etal \cite{zhou2019davanet}, which contains 20,637 blurry-sharp stereo image pairs from 126 diverse sequences and their corresponding bidirectional disparities.
To reliably synthesize events, we first increase the sequence frame rate from 60 fps to 2,400 fps via a high-quality frame interpolation algorithm \cite{jiang2018super} and then applying the V2E event simulator \cite{delbruck2020v2e} to the high frame rate sequences.
We use the officially split method, where 89 sequences are used for self-supervised training, and 37 sequences are used for testing.
For the real sensor data, we use the MVSEC \cite{zhu2018multivehicle} dataset, which contains the stereo intensity images and events captured by DAVIS 240C.
MVSEC also provides ground truth depth captured by LiDAR.
We use the officially split method for the MVSEC dataset.
Our method is implemented using Pytorch \cite{paszke2019pytorch} framework and trained using NVIDIA V100 GPUs.
For the stereo network design, we build our multi-modal networks by modifying DeepPruner \cite{duggal2019deeppruner} and AANet \cite{xu2020aanet} according to Sec. \ref{sec:method:stereo}.
Note that our framework is compatible with the most alternative architectures of the reconstruction and the stereo networks.
For optimization, we use Adam \cite{kingma2015adam} with $\beta_1 = 0.9$, $\beta_2 = 0.999$ and learning rate $1 \times 10^{-4}$.
We set the weighting of the different loss components to $\lambda_{gd}=1$, $\lambda_{sm}=0.1$,  $\lambda_{cc}=0.025$ and $\lambda_{itn}=0.005$.
The settings of $\rho$ and $t$ are experimental. The final value of $\rho$ is 1 on the synthetic dataset and 1.5 on the real world data. The value of $t$ is 2 for all datasets.
The overall self-supervised training costs about 2 days.

\begin{figure}[t]
    \centering
    \includegraphics[width=\linewidth]{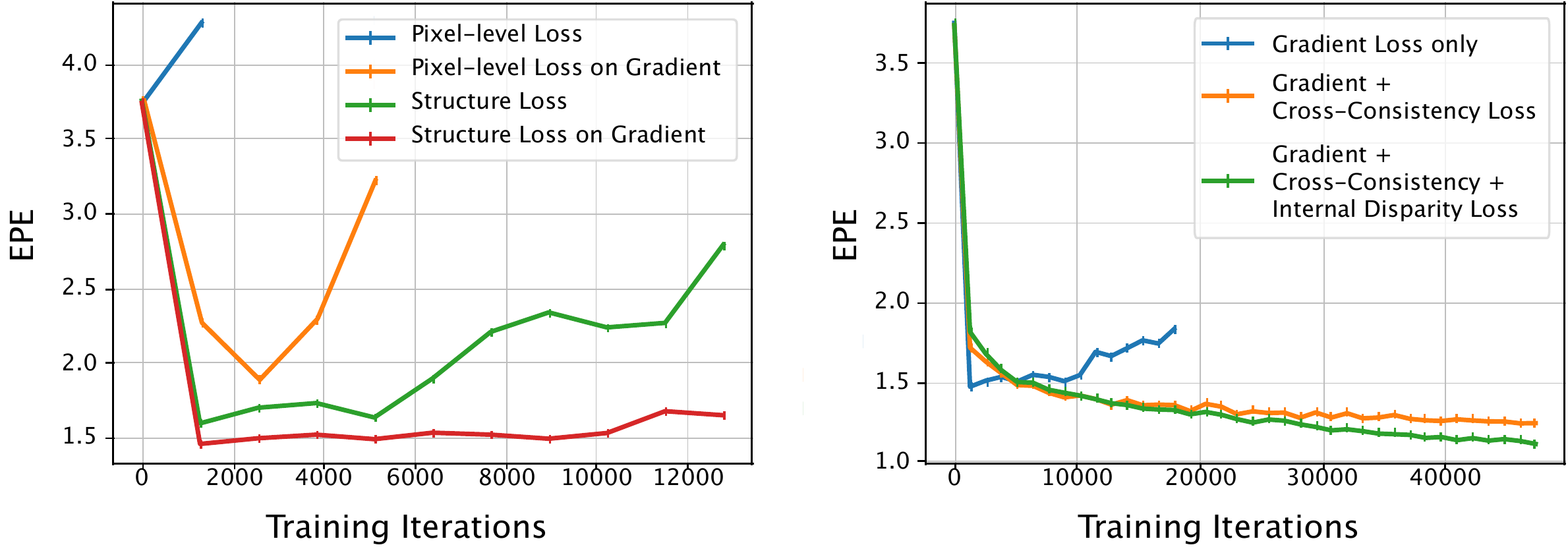}
    \caption{The convergence curves of using different loss. The left figure shows that the proposed gradient structure loss can provide meaningful self-supervision. The right figure indicates that the proposed general stereo loss could produce robust training effects. Some methods failed in the middle of training.}
    \label{fig:loss_comp}
\end{figure}

\subsection{Stereo Matching Results}
\label{sec:exp:matching}
We first quantitatively demonstrate the effectiveness of the proposed method in \tablename \ref{tab:comparison} and \tablename \ref{tab:comparison_real}.
The metrics are averaged end-point error (EPE) and $>$1-pixel, $>$3-pixel and $>$5-pixel error.
Under the proposed setting, only limited methods can be used to obtain disparity maps as there is no multi-modal stereo matching model for event-intensity setting.
We included the monocular depth model \cite{godard2019digging} for comparison, which was not fine-tuned on the target data, as we cannot obtain intensity images from both left and right views at the same time in this setting theoretically.
We also consider the monocular event depth model \cite{gehrig2021combining}, but the advantage of the event model is to combat high-speed motion and low-light situations that the intensity camera cannot handle, and its effect cannot be compared with the results predicted using the intensity images.
We can have the following observations.
Firstly, the proposed method can achieve much better results compared with monocular disparity estimation results using monodepthv2 \cite{godard2019digging} that is only based on the left image.
It indicates that the information of another view plays an essential role in stereo matching.
The unsatisfactory effect prevents us from using this disparity map to align events with the intensity image.
Secondly, we show the results of directly performing stereo matching using the reconstructed right image and left image (marked as the ``baseline experiments'').
As can be seen, since the reconstructed images have a huge color difference against the left view image, it isn't easy to obtain a good result by employing the existing stereo vision models.
The visualization results in \figurename \ref{fig:main_synthetic} also speak to similar conclusion.
However, introducing the modified stereo network and self-supervised learning using only gradient structure loss can improve the stereo estimation results.
Even if all the loss functions are used, the network architecture is still a significant obstacle to improving performance.
Thirdly, the proposed self-supervised learning method unleashes the full potential of the overall framework's effect, which proves the advantage of our self-supervised learning strategy, that is, learning from unlabeled data.
We visualize the results of the proposed method in \figurename \ref{fig:main_synthetic} and \figurename \ref{fig:comp_mvsec}.
As one can observe, we can only get poor matching results in these scenarios based on the pre-trained stereo network.
Our full model produces accurate object boundaries and better preserves the overall scene structures.
We also provide the upper bound performance obtained by a fine-tuned DeepPruner network using both sides' intensity images as a reference.
It can be seen that the information lost by the event is detrimental to the final matching result.
But the purpose of our method is not to rely on events to obtain better matching results but to make it possible to calculate reasonable disparity under the proposed intensity-event setting.

\subsection{Ablation Study}
\label{sec:exp:ablation}
\begin{figure}[t]
    \centering
        \includegraphics[width=0.7\textwidth]{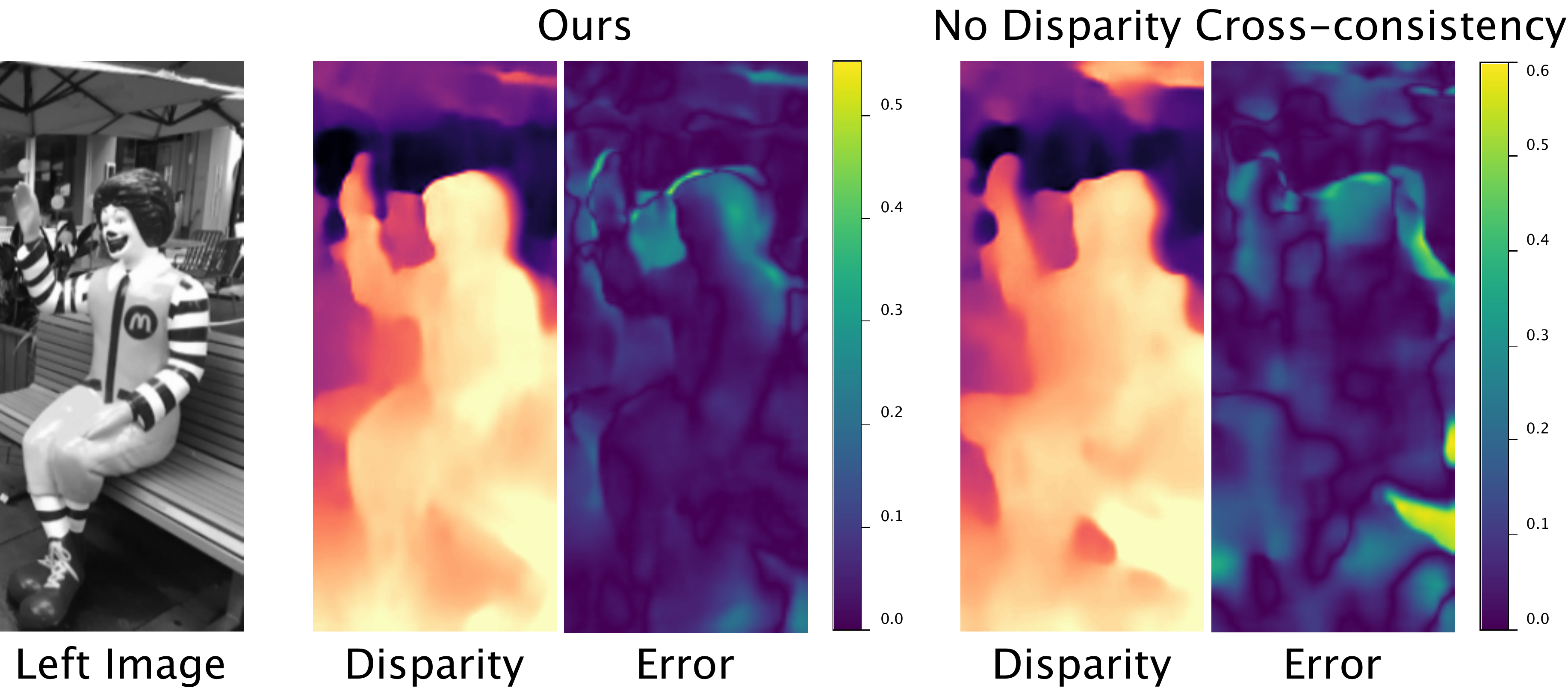}
        \caption{Comparison between our method with and without the cross-consistency loss. Our consistency term helps outline the precise edge in the disparity map.}
    \label{fig:disploss}
\end{figure}

To study the effects of each component in the proposed method, we conduct several ablation studies.
All the experiments are conducted using FireNet reconstruction and a modified AANet stereo model.
We first examine the use of gradient structure loss function.
We train our model using the four alternative loss functions described in Sec \ref{sec:method:ssl} respectively, and their convergence curves are shown in the left figure of \figurename \ref{fig:loss_comp}.
As can be seen, all pixel-wise loss functions fail to converge.
They can only provide very limited information, and continuous optimization of these losses will bring adverse effects and make training fail.
The structure loss directly calculated on images performs well initially, but it could no longer provide adequate supervision as the training progressed.
The proposed gradient structural loss can produce a relatively reliable convergence curve, although it is still tricky to improve performance with training steadily.
We next involve the proposed cross-consistency loss and internal disparity loss in training.
The results are shown in right figure of \figurename \ref{fig:loss_comp}.
It is surprising that although it cannot be compared with competitors initially, the proposed losses make the training process more stable.
The proposed method can steadily and continuously improve performance without any ground truth data.
We visually explain why the proposed cross-consistency loss is so effective.
\figurename \ref{fig:disploss} shows some comparisons between models trained with and without the cross-consistency loss.
The results speak to the fact that the inclusion of this loss improves the quality of the result.
This loss term helps outline the precise edge and generate sharp, accurate shapes in the disparity map.

\subsection{Connecting Events and Intensity}
\label{sec:exp:app}
\begin{figure}[t]
    \centering
    \includegraphics[width=\linewidth]{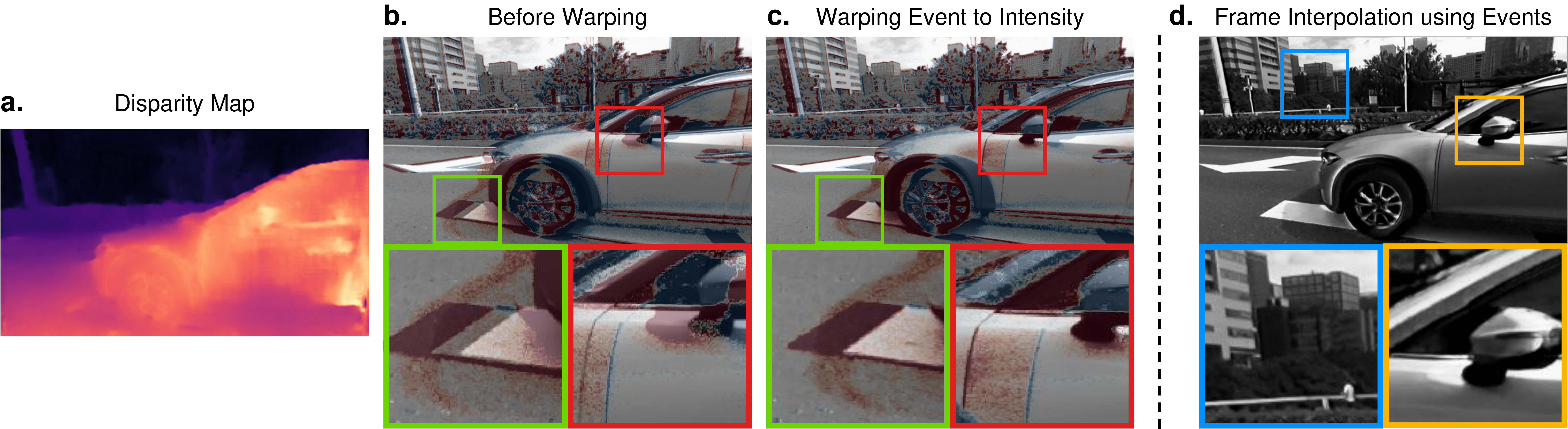}
    \caption{Connecting events and intensity. {$\textbf{a}$}, the calculated disparity map using the proposed method. {$\textbf{b}$}, before warping, events and intensity image are not aligned spatially. {$\textbf{c}$}, after warping using disparity map, the events and intensity image are well aligned. {$\textbf{d}$}, we employ the warped events and intensity image to perform temporal frame interpolation using \cite{lin2020learning}.}
    \label{fig:event_warp}
\end{figure}

We next demonstrate the use of a disparity map to establish a connection between the intensity camera and the event camera.
Due to the displacement between these two cameras, even if we can obtain the left view intensity image and the right view events, the disparity between them makes it difficult to make full use of the advantages of the two sensors, as shown in \figurename \ref{fig:event_warp} (b).
In this case, many algorithms and applications that require alignment of events and images cannot be implemented, \eg, \cite{duan2021guided,lin2020learning}.
We first obtain the disparity map using the proposed multi-modal stereo method.
Each value in the disparity map indicates the number of pixels that need to be shifted horizontally.
We warp events by changing the x coordinate in each event tuple $(x_m,y_m,t_m,p_m)$, where $x_m$, $y_m$, $t_m$ denote the spatial-temporal coordinates, and $p_m\in\{-1,+1\}$ denotes the polarity of the event.
The warped event are visualized in \figurename \ref{fig:event_warp} (c).
It can be seen that the warped right view events and the left view image are well aligned both spatially and temporally.
Obtaining such a connection between two sensors allows many downstream tasks.
Here we show the application potential by the event-based video interpolation task.
Motivated by the physical model of event that the residuals between a blurry image and sharp frames are the integrals of events, Lin \etal \cite{lin2020learning} propose to estimate the residuals for the sharp frame restoration based on events.
Our reconstruction result shows good fidelity performance, which further proves the application value of the proposed problem setting.

\subsection{Limitations}
\label{sec:exp:failure}
\begin{figure}[t]
    \centering
        \includegraphics[width=0.8\textwidth]{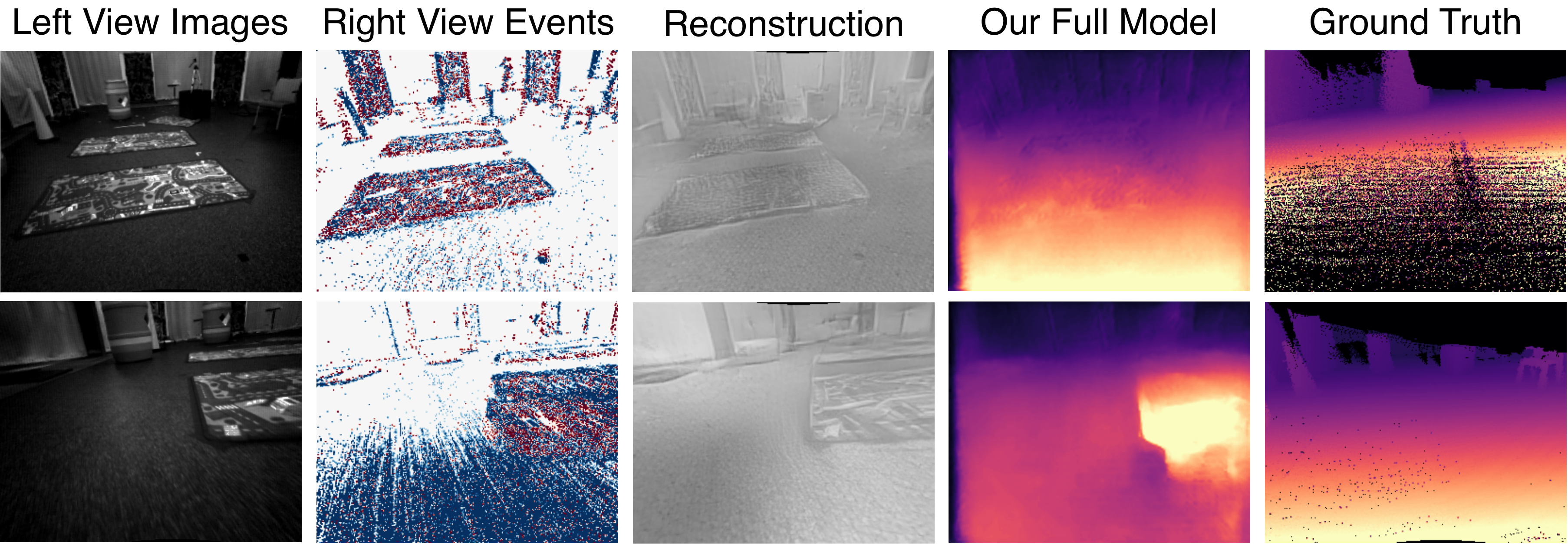}
        \caption{Failure cases visualization. The proposed method faces challenges when the reconstruction only contains very limited information.}
    \label{fig:fail}
\end{figure}

At last, we show some failure cases and analyze the potential limitations.
We show two failure cases in \figurename \ref{fig:fail} and both of them are from MVSEC.
The direct reconstruction using E2VID contains only very limited information, resulting in the failure of stereo matching.
This shows a possible flaw of the proposed method, that is, it still faces significant challenges when reconstruction results are very vague.
A possible solution is to design a stereo network to perform stereo matching between event streams and intensity images directly.
In that case, the loss functions described in this paper can still provide good self-supervised learning results.

\section{Conclusion}
\label{sec:con}
This paper presents a novel camera setting with an intensity camera and an event camera and establishes a connection between them with a multi-modal stereo matching task.
Based on the proposed self-supervised method, we can obtain fine disparity maps under this novel setting and not collect any ground truth disparities.
Experiments demonstrate the effectiveness and the application value of the proposed method.


\bibliography{report}   

\begin{thebibliography}{10}

\bibitem{rebecq2016evo}
H.~Rebecq, T.~Horstsch{\"a}fer, G.~Gallego, {\em et~al.}, ``Evo: A geometric
  approach to event-based 6-dof parallel tracking and mapping in real time,''
  {\em IEEE Robotics and Automation Letters} {\bf 2}(2), 593--600  (2016).

\bibitem{jiang2020learning}
Z.~Jiang, Y.~Zhang, D.~Zou, {\em et~al.}, ``Learning event-based motion
  deblurring,'' in {\em CVPR},  3320--3329  (2020).

\bibitem{zhu2018ev}
A.~Z. Zhu and L.~Yuan, ``Ev-flownet: Self-supervised optical flow estimation
  for event-based cameras,'' in {\em Robotics: Science and Systems},   (2018).

\bibitem{vidal2018ultimate}
A.~R. Vidal, H.~Rebecq, T.~Horstschaefer, {\em et~al.}, ``Ultimate slam?
  combining events, images, and imu for robust visual slam in hdr and
  high-speed scenarios,'' {\em IEEE Robotics and Automation Letters} {\bf
  3}(2), 994--1001  (2018).

\bibitem{tulyakov2021time}
S.~Tulyakov, D.~Gehrig, S.~Georgoulis, {\em et~al.}, ``Time lens: Event-based
  video frame interpolation,'' in {\em CVPR},  16155--16164  (2021).

\bibitem{patrick2008128x}
L.~Patrick, C.~Posch, and T.~Delbruck, ``A 128x 128 120 db 15$\mu$ s latency
  asynchronous temporal contrast vision sensor,'' {\em IEEE Journal of
  Solid-State Circuits} {\bf 43}, 566--576  (2008).

\bibitem{brandli2014240}
C.~Brandli, R.~Berner, M.~Yang, {\em et~al.}, ``A 240$\times$ 180 130 db 3
  $\mu$s latency global shutter spatiotemporal vision sensor,'' {\em IEEE
  Journal of Solid-State Circuits} {\bf 49}(10), 2333--2341  (2014).

\bibitem{rebecq2019events}
H.~Rebecq, R.~Ranftl, V.~Koltun, {\em et~al.}, ``Events-to-video: Bringing
  modern computer vision to event cameras,'' in {\em CVPR},  3857--3866
  (2019).

\bibitem{scheerlinck2020fast}
C.~Scheerlinck, H.~Rebecq, D.~Gehrig, {\em et~al.}, ``Fast image reconstruction
  with an event camera,'' in {\em WACV},  156--163  (2020).

\bibitem{gehrig2018asynchronous}
D.~Gehrig, H.~Rebecq, G.~Gallego, {\em et~al.}, ``Asynchronous, photometric
  feature tracking using events and frames,'' in {\em ECCV},  750--765  (2018).

\bibitem{lin2020learning}
S.~Lin, J.~Zhang, J.~Pan, {\em et~al.}, ``Learning event-driven video
  deblurring and interpolation,'' in {\em ECCV},  695--710  (2020).

\bibitem{zhu2019unsupervised}
A.~Z. Zhu, L.~Yuan, K.~Chaney, {\em et~al.}, ``Unsupervised event-based
  learning of optical flow, depth, and egomotion,'' in {\em CVPR},   (2019).

\bibitem{kim2016real}
H.~Kim, S.~Leutenegger, and A.~J. Davison, ``Real-time 3d reconstruction and
  6-dof tracking with an event camera,'' in {\em ECCV},  349--364  (2016).

\bibitem{kueng2016low}
B.~Kueng, E.~Mueggler, G.~Gallego, {\em et~al.}, ``Low-latency visual odometry
  using event-based feature tracks,'' in {\em IROS},  16--23, IEEE  (2016).

\bibitem{barua-event2img}
S.~Barua, Y.~Miyatani, and A.~Veeraraghavan, ``Direct face detection and video
  reconstruction from event cameras,'' {\em WACV}   (2016).

\bibitem{bardow}
P.~Bardow, A.~J. Davison, and S.~Leutenegger, ``Simultaneous optical flow and
  intensity estimation from an event camera,'' {\em CVPR}   (2016).

\bibitem{mostafavi_cgan_event_image_translation}
S.~M. Mostafavi~I., L.~Wang, Y.~Ho, {\em et~al.}, ``Event-based high dynamic
  range image and very high frame rate video generation using conditional
  generative adversarial networks,'' {\em CVPR}   (2019).

\bibitem{mostafavi2020e2sri}
S.~M. Mostafavi~I., J.~Choi, and K.-J. Yoon, ``Learning to super resolve
  intensity images from events,'' {\em CVPR}   (2020).

\bibitem{Wang_2020_CVPR_eventSR}
L.~Wang, T.-K. Kim, and K.-J. Yoon, ``Eventsr: From asynchronous events to
  image reconstruction, restoration, and super-resolution via end-to-end
  adversarial learning,'' in {\em CVPR},   (2020).

\bibitem{stoffregen2020reducing}
T.~Stoffregen, C.~Scheerlinck, D.~Scaramuzza, {\em et~al.}, ``Reducing the
  sim-to-real gap for event cameras,'' in {\em ECCV},   (2020).

\bibitem{Yoon_stereo_early}
K.-J. Yoon and I.~S. Kweon, ``Adaptive support-weight approach for
  correspondence search,'' {\em IEEE TPAMI}   (2006).

\bibitem{hosni_stereo_early}
A.~Hosni, C.~Rhemann, M.~Bleyer, {\em et~al.}, ``Fast cost-volume filtering for
  visual correspondence and beyond,'' {\em IEEE TPAMI}   (2013).

\bibitem{Zbontar_2015_CVPR}
J.~Zbontar and Y.~LeCun, ``Computing the stereo matching cost with a
  convolutional neural network,'' in {\em CVPR},   (2015).

\bibitem{luo2016efficient}
W.~Luo, A.~G. Schwing, and R.~Urtasun, ``Efficient deep learning for stereo
  matching,'' in {\em CVPR},  5695--5703  (2016).

\bibitem{mayer2016large}
N.~Mayer, E.~Ilg, P.~Hausser, {\em et~al.}, ``A large dataset to train
  convolutional networks for disparity, optical flow, and scene flow
  estimation,'' in {\em CVPR},  4040--4048  (2016).

\bibitem{Kendall_2017_ICCV}
A.~Kendall, H.~Martirosyan, S.~Dasgupta, {\em et~al.}, ``End-to-end learning of
  geometry and context for deep stereo regression,'' in {\em ICCV},   (2017).

\bibitem{Chang_2018_CVPR}
J.-R. Chang and Y.-S. Chen, ``Pyramid stereo matching network,'' in {\em CVPR},
    (2018).

\bibitem{Zhang_2019_GANet}
F.~Zhang, V.~Prisacariu, R.~Yang, {\em et~al.}, ``Ga-net: Guided aggregation
  net for end-to-end stereo matching,'' in {\em CVPR},   (2019).

\bibitem{duggal2019deeppruner}
S.~Duggal, S.~Wang, W.-C. Ma, {\em et~al.}, ``Deeppruner: Learning efficient
  stereo matching via differentiable patchmatch,'' in {\em ICCV},  4384--4393
  (2019).

\bibitem{xu2020aanet}
H.~Xu and J.~Zhang, ``Aanet: Adaptive aggregation network for efficient stereo
  matching,'' in {\em CVPR},  1959--1968  (2020).

\bibitem{chiu2011improving}
W.~W.-C. Chiu, U.~Blanke, and M.~Fritz, ``Improving the kinect by cross-modal
  stereo.,'' in {\em BMVC},  116.1--116.10  (2011).

\bibitem{zhi2018deep}
T.~Zhi, B.~R. Pires, M.~Hebert, {\em et~al.}, ``Deep material-aware
  cross-spectral stereo matching,'' in {\em Proceedings of the IEEE Conference
  on Computer Vision and Pattern Recognition},  1916--1925  (2018).

\bibitem{shen2014multi}
X.~Shen, L.~Xu, Q.~Zhang, {\em et~al.}, ``Multi-modal and multi-spectral
  registration for natural images,'' in {\em ECCV},  309--324  (2014).

\bibitem{jeon2016stereo}
H.-G. Jeon, J.-Y. Lee, S.~Im, {\em et~al.}, ``Stereo matching with color and
  monochrome cameras in low-light conditions,'' in {\em CVPR},  4086--4094
  (2016).

\bibitem{kim2015dasc}
S.~Kim, D.~Min, B.~Ham, {\em et~al.}, ``Dasc: Dense adaptive self-correlation
  descriptor for multi-modal and multi-spectral correspondence,'' in {\em
  CVPR},  2103--2112  (2015).

\bibitem{kim2016deep}
S.~Kim, D.~Min, S.~Lin, {\em et~al.}, ``Deep self-correlation descriptor for
  dense cross-modal correspondence,'' in {\em European Conference on Computer
  Vision},  679--695, Springer  (2016).

\bibitem{Mostafavi_2021_ICCV}
M.~Mostafavi, K.-J. Yoon, and J.~Choi, ``Event-intensity stereo: Estimating
  depth by the best of both worlds,'' in {\em ICCV},  4258--4267  (2021).

\bibitem{garg2016unsupervised}
R.~Garg, B.~V. Kumar, G.~Carneiro, {\em et~al.}, ``Unsupervised cnn for single
  view depth estimation: Geometry to the rescue,'' in {\em ECCV},  740--756
  (2016).

\bibitem{godard2017unsupervised}
C.~Godard, O.~Mac~Aodha, and G.~J. Brostow, ``Unsupervised monocular depth
  estimation with left-right consistency,'' in {\em CVPR},   (2017).

\bibitem{godard2019digging}
C.~Godard, O.~Mac~Aodha, M.~Firman, {\em et~al.}, ``Digging into
  self-supervised monocular depth estimation,'' in {\em ICCV},  3828--3838
  (2019).

\bibitem{zhou2017unsupervised}
C.~Zhou, H.~Zhang, X.~Shen, {\em et~al.}, ``Unsupervised learning of stereo
  matching,'' in {\em ICCV},  1567--1575  (2017).

\bibitem{rebecq2019high}
H.~Rebecq, R.~Ranftl, V.~Koltun, {\em et~al.}, ``High speed and high dynamic
  range video with an event camera,'' {\em IEEE TPAMI} {\bf 43}(6), 1964--1980
  (2021).

\bibitem{zhu2018multivehicle}
A.~Z. Zhu, D.~Thakur, T.~{\"O}zaslan, {\em et~al.}, ``The multivehicle stereo
  event camera dataset: An event camera dataset for 3d perception,'' {\em IEEE
  Robotics and Automation Letters} {\bf 3}(3), 2032--2039  (2018).

\bibitem{zhang2018unreasonable}
R.~Zhang, P.~Isola, A.~A. Efros, {\em et~al.}, ``The unreasonable effectiveness
  of deep features as a perceptual metric,'' in {\em CVPR},  586--595  (2018).

\bibitem{jinjin2020pipal}
G.~Jinjin, C.~Haoming, C.~Haoyu, {\em et~al.}, ``Pipal: a large-scale image
  quality assessment dataset for perceptual image restoration,'' in {\em ECCV},
   633--651  (2020).

\bibitem{wang2004image}
Z.~Wang, A.~C. Bovik, H.~R. Sheikh, {\em et~al.}, ``Image quality assessment:
  from error visibility to structural similarity,'' {\em IEEE transactions on
  image processing} {\bf 13}(4), 600--612, IEEE  (2004).

\bibitem{zhou2019davanet}
S.~Zhou, J.~Zhang, W.~Zuo, {\em et~al.}, ``Davanet: Stereo deblurring with view
  aggregation,'' in {\em CVPR},  10996--11005  (2019).

\bibitem{jiang2018super}
H.~Jiang, D.~Sun, V.~Jampani, {\em et~al.}, ``Super slomo: High quality
  estimation of multiple intermediate frames for video interpolation,'' in {\em
  CVPR},  9000--9008  (2018).

\bibitem{delbruck2020v2e}
T.~Delbruck, Y.~Hu, and Z.~He, ``{V2E}: From video frames to realistic {DVS}
  event camera streams,'' in {\em CVPRW},   (2021).

\bibitem{paszke2019pytorch}
A.~Paszke, S.~Gross, F.~Massa, {\em et~al.}, ``Pytorch: An imperative style,
  high-performance deep learning library,'' {\em NeurIPS} , 8026--8037  (2019).

\bibitem{kingma2015adam}
D.~P. Kingma and J.~Ba, ``Adam: A method for stochastic optimization,'' in {\em
  ICLR},   (2015).

\bibitem{gehrig2021combining}
D.~Gehrig, M.~R{\"u}egg, M.~Gehrig, {\em et~al.}, ``Combining events and frames
  using recurrent asynchronous multimodal networks for monocular depth
  prediction,'' {\em IEEE Robotics and Automation Letters} {\bf 6}(2),
  2822--2829  (2021).

\bibitem{duan2021guided}
P.~Duan, Z.~Wang, B.~Shi, {\em et~al.}, ``Guided event filtering: Synergy
  between intensity images and neuromorphic events for high performance
  imaging,'' {\em IEEE Transactions on Pattern Analysis and Machine
  Intelligence}   (2021).

\bibitem{wang2018esrgan}
X.~Wang, K.~Yu, S.~Wu, {\em et~al.}, ``Esrgan: Enhanced super-resolution
  generative adversarial networks,'' in {\em ECCV},  63--79  (2018).

\end{thebibliography}
\bibliographystyle{spiejour}   

\appendix
\section{More Stereo Matching Results}
\label{sec:apd:stereo}
\begin{figure} 
  \centering 
  \includegraphics[width=\linewidth]{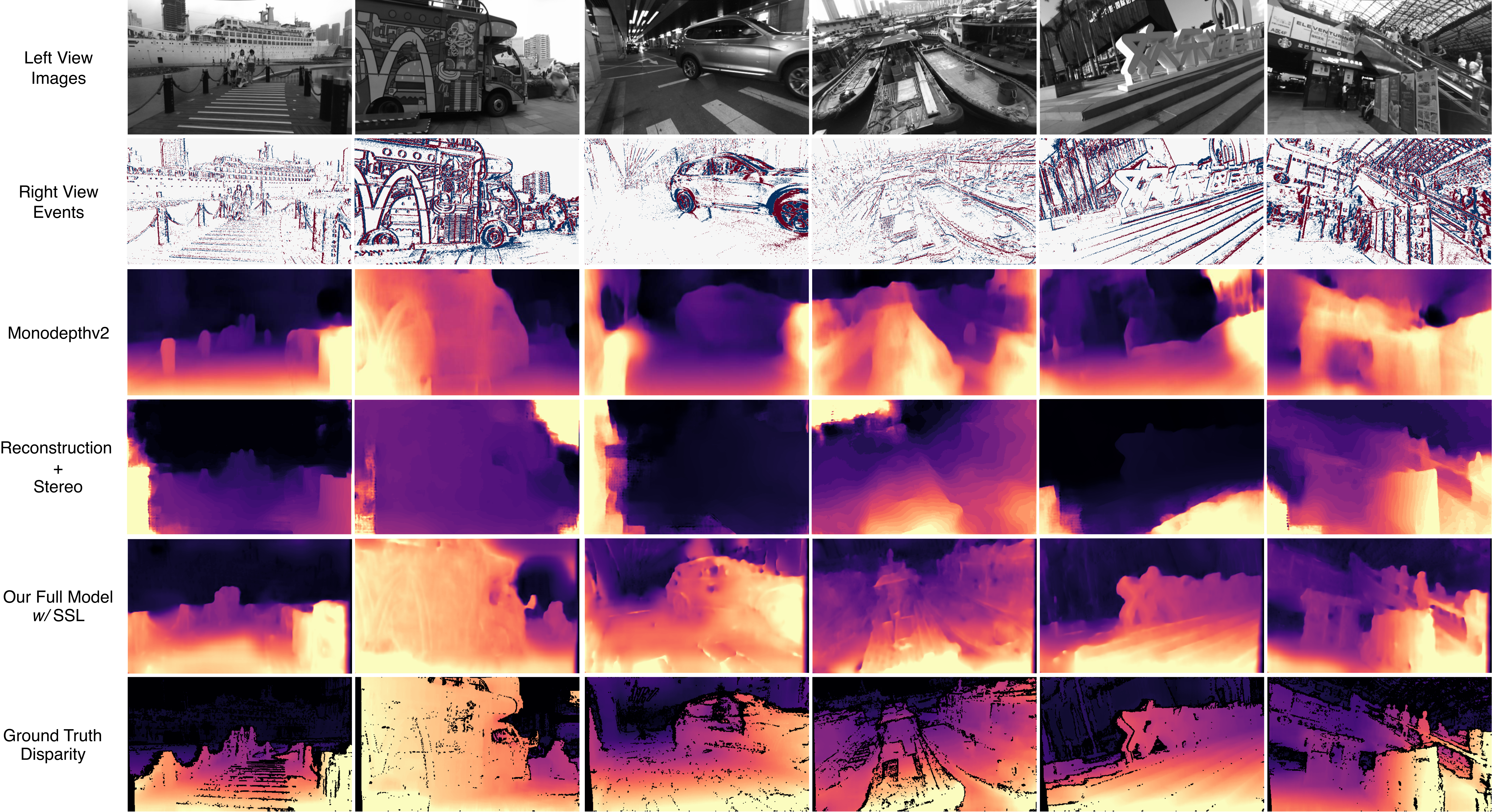}
  \caption{Stereo matching results of different methods on synthetic dataset.}
  \label{fig:apd:stereo_main}
\end{figure}

\begin{figure} 
  \centering 
  \includegraphics[width=\linewidth]{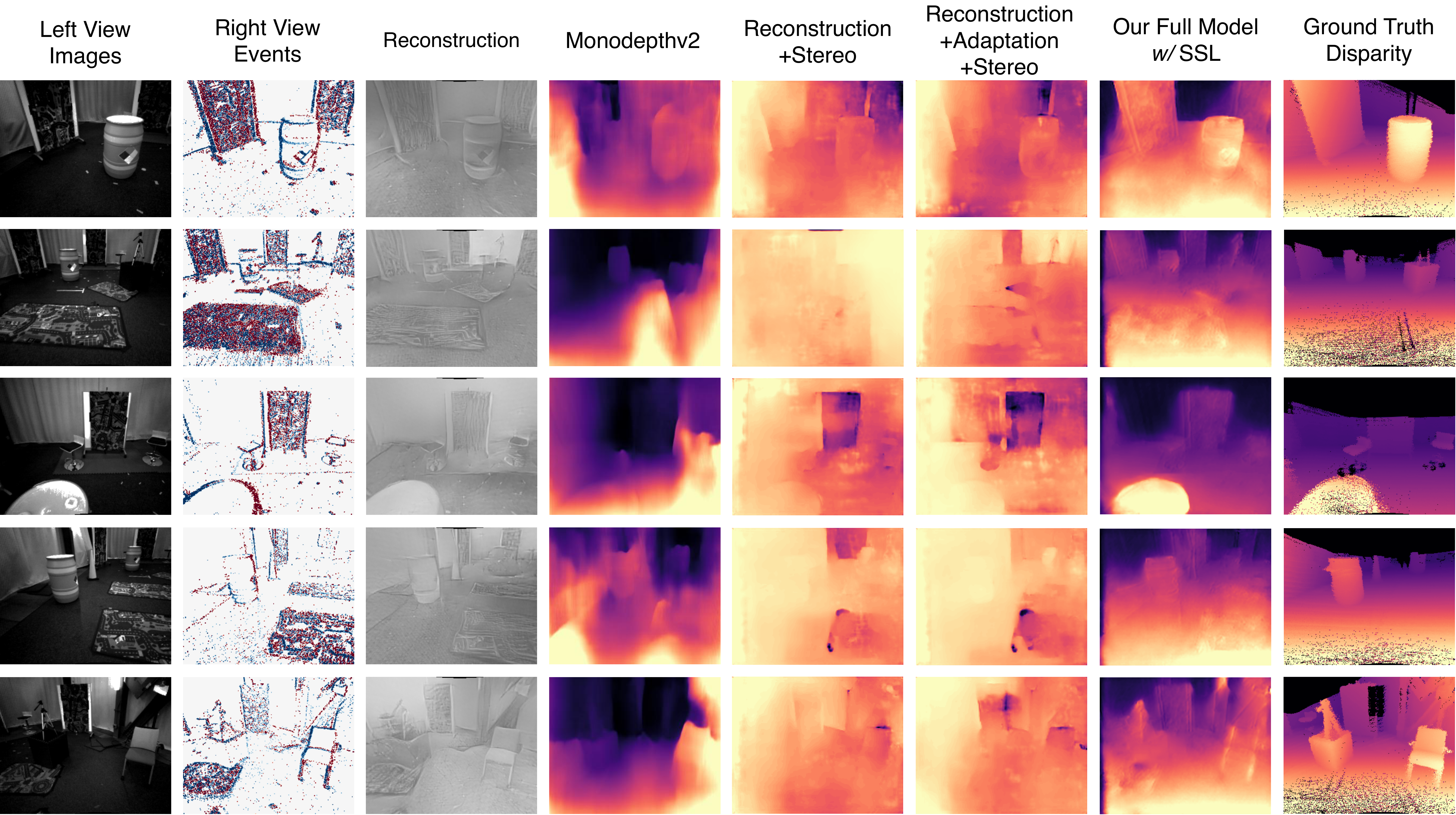}
  \caption{Stereo matching comparison of different methods on the MVSEC \cite{zhu2018multivehicle} real-world dataset.}
  \label{fig:apd:stereo_mvsec}
\end{figure}

We first provide more stereo matching results using different methods.
In \figurename~\ref{fig:apd:stereo_main}, we provide the comparison on synthetic events data.
The synthesis method is described in Sec 4.1 in the main text.
In \figurename~\ref{fig:apd:stereo_mvsec}, we provide the comparison results on the MVSEC \cite{zhu2018multivehicle} dataset.
In these experiments, we use E2VID as the reconstruction network and AANet \cite{xu2020aanet} as the stereo matching network.
It can be seen that our method provides the best results and performs well in these cases.
In \figurename~\ref{fig:apd:stereo_mvsec}, we also compare our method with another commonly used method when encountering multi-modal problems, \eg, Zhi \etal \cite{zhi2018deep} propose to use a spectral translation network to facilitate cross-spectral stereo matching.
We develop a similar adaptation network to translate the rough reconstructed right view image to add color information.
The network structure is ResNet image translation structure, similar to SRResNet \cite{wang2018esrgan} but without an upsampling layer.
We use a supervised training method to train this network, and the Ground Truth of the intensity image is provided in MVSEC.
As can be seen, simply using the modality alignment method cannot bring better results. There are two main reasons for this.
Firstly, the color information has been lost in the events and reconstruction results and can not be simply recovered by an adaptation network.
Secondly, the pre-trained stereo model can not generalize well in the MVSEC dataset.
This also provides the necessities of learning using target data in a self-supervised manner.

\begin{figure*}
    \centering
    \includegraphics[width=0.45\linewidth]{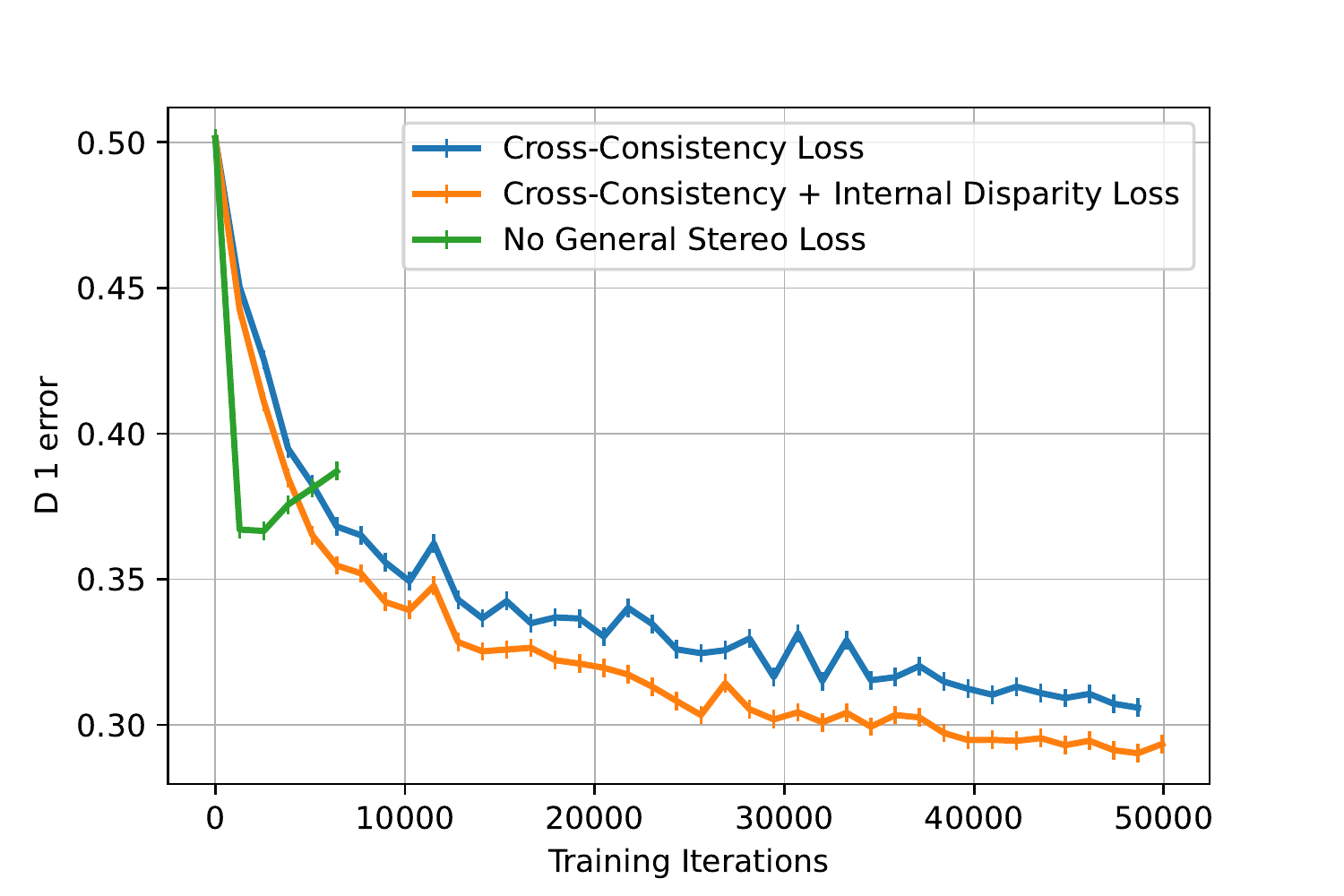}
    \includegraphics[width=0.45\linewidth]{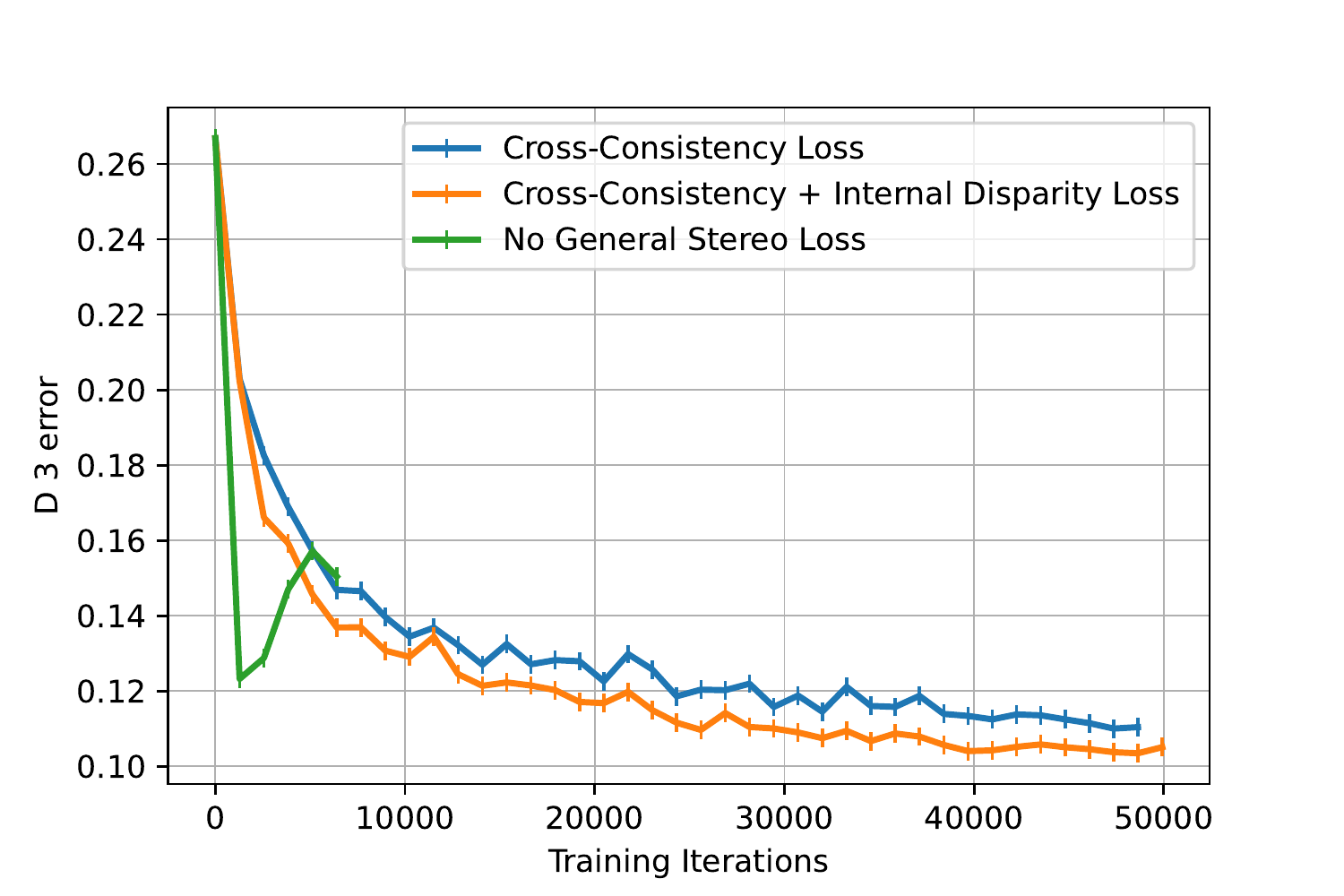}
    \includegraphics[width=0.45\linewidth]{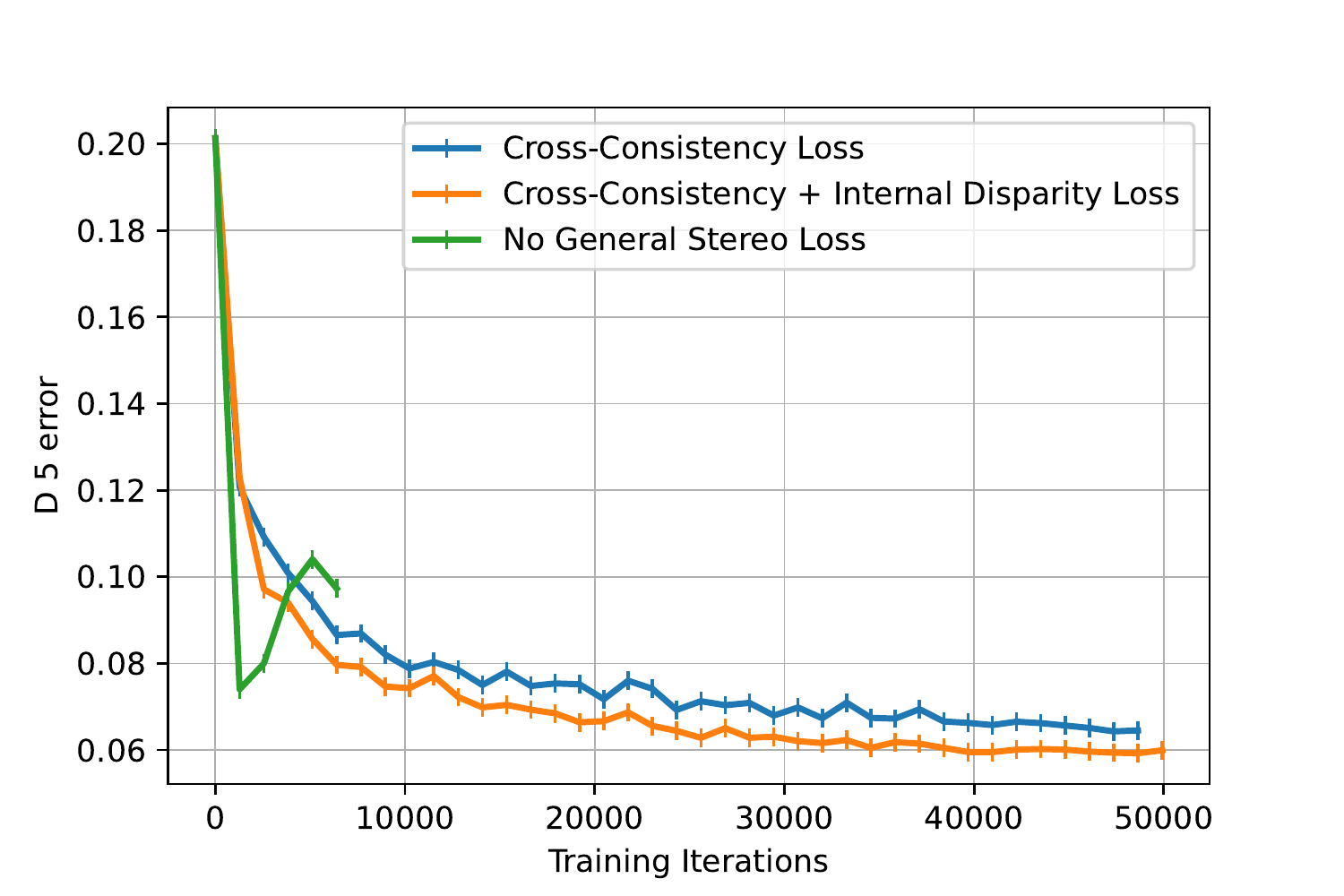}
    \includegraphics[width=0.45\linewidth]{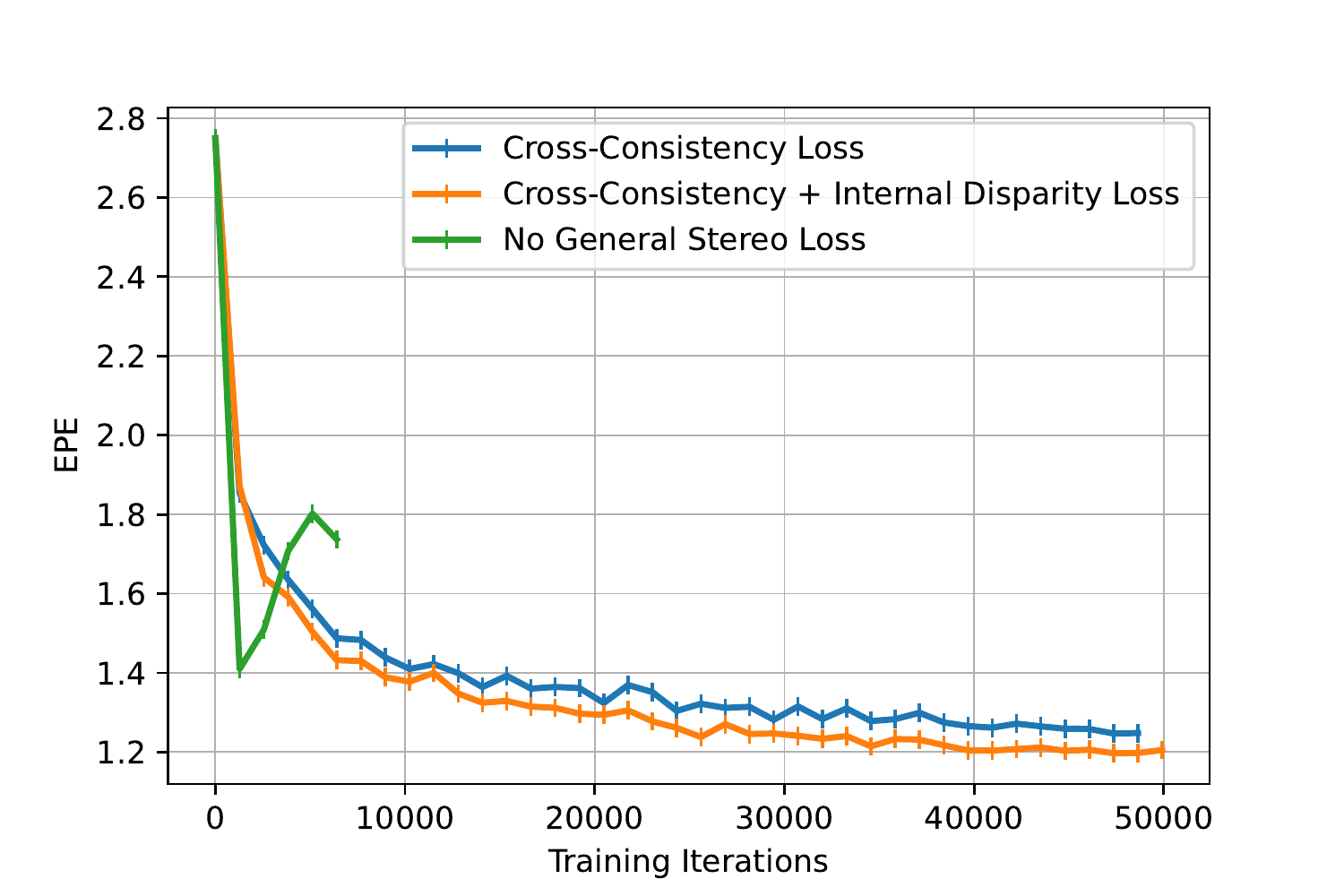}
    \caption{The convergence curves of using different losses. The proposed general stereo loss could produce robust training effects. The method of ``no general stereo losses'' failed in the middle of training.}
    \label{fig:loss_curves}
\end{figure*}

\section{Event Warping and Video Frame Interpolation}
\label{sec:apd:warp}
\begin{figure}
    \centering
    \includegraphics[width=\linewidth]{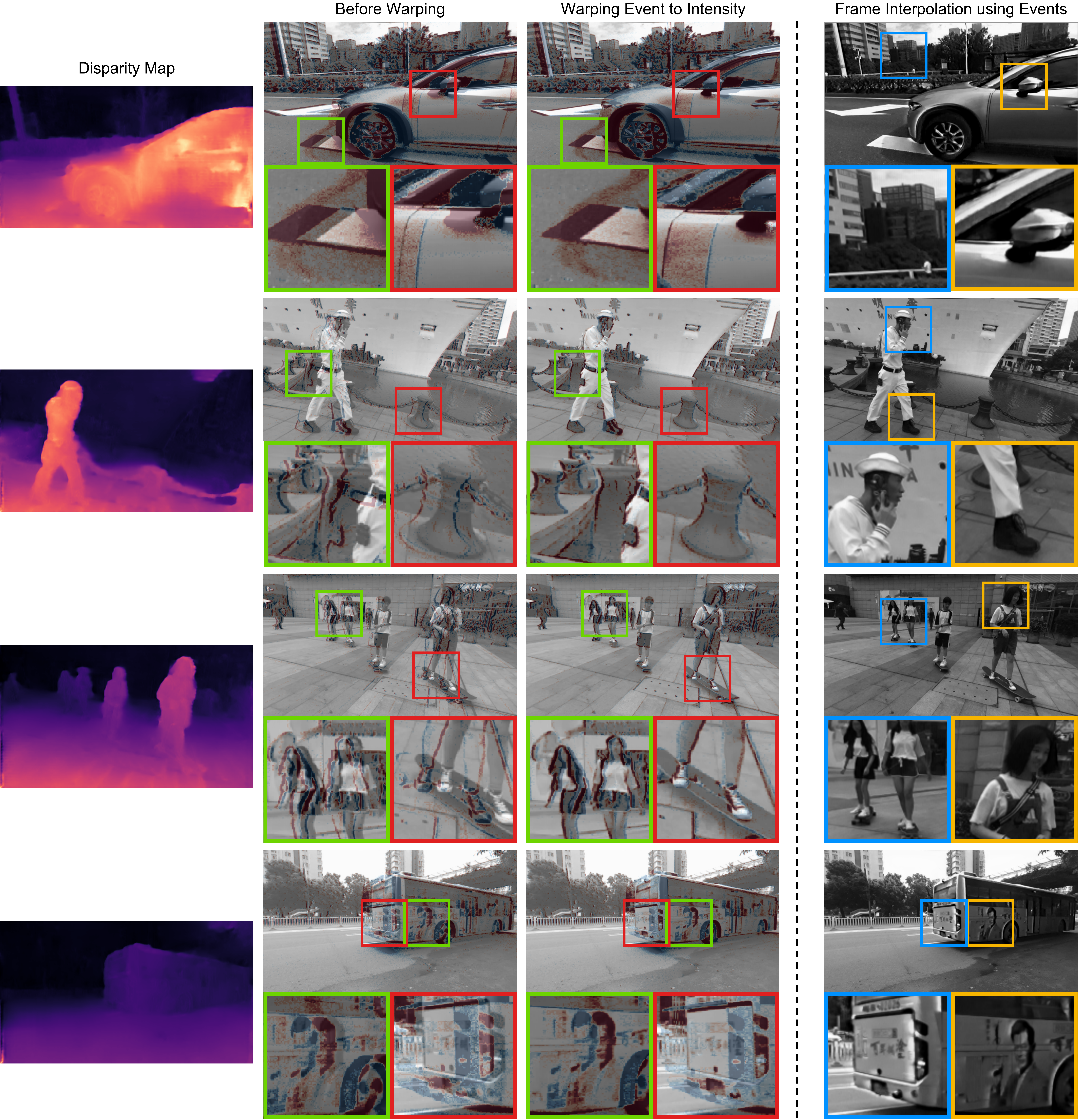}
    \caption{Connecting events and intensity. The event warping results and the temporal frame interpolation results using the warped events and intensity images using \cite{lin2020learning}.}
    \label{fig:apd:warp}
\end{figure}

In this section, we show more results of warping events to intensity image.
We first obtain the disparity between these two sensors using the proposed method.
The warping provides us with events and intensity images that are aligned both spatially and temporally.
We can use the obtained events and images to support downstream applications.
In this supplementary material, we show more video frame interpolation \cite{lin2020learning} results in \figurename~\ref{fig:apd:warp}.

The proposed method has additional value in this respect.
Due to the hardware limitations of the event camera, high-resolution events and high-resolution intensity images are not available simultaneously.
But with the proposed method, we can obtain high-resolution events and images simultaneously through two sensors.
This makes a series of applications, such as video frame interpolation possible.

\section{Effect of the Loss functions}
\label{sec:apd:loss}
\begin{figure}
    \centering
    \includegraphics[width=\linewidth]{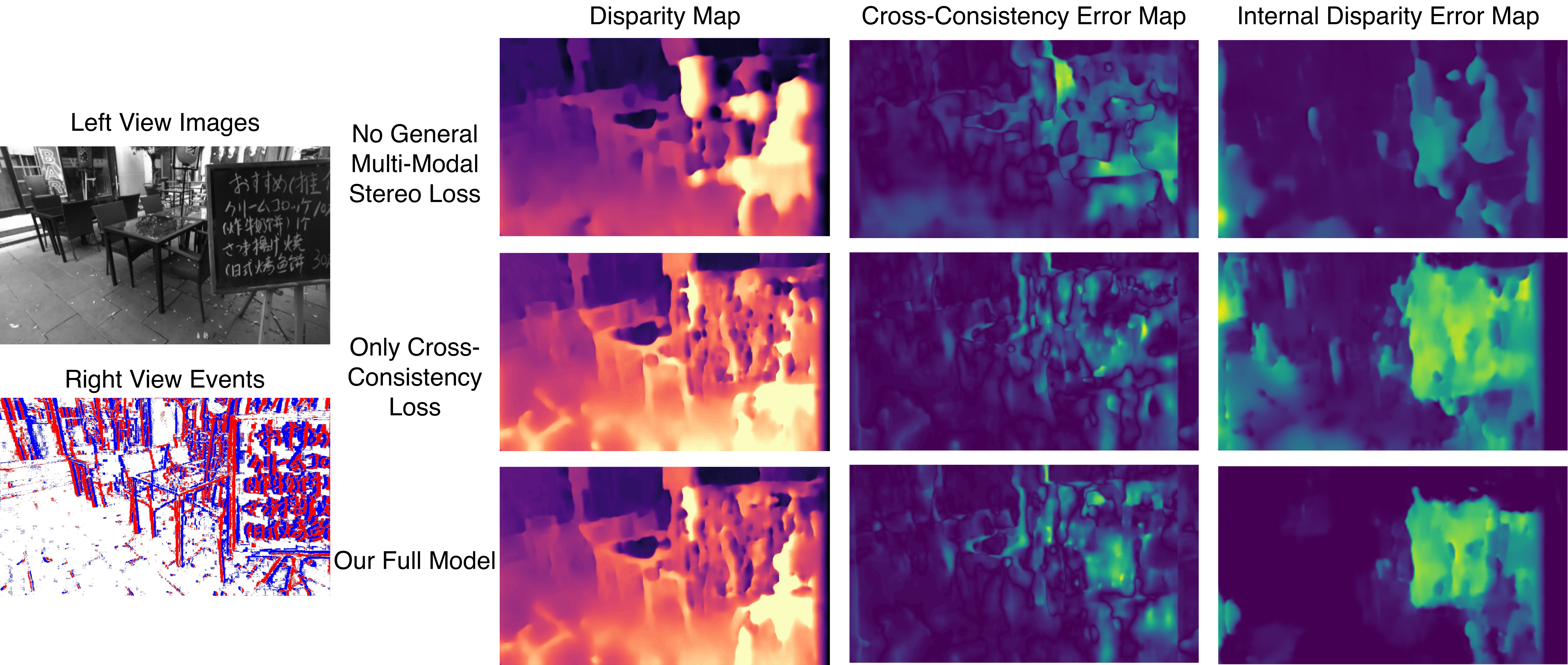}
    \caption{Visualization results of the proposed loss functions. This is a failure case. The disparity of the front object exceeds the upper limit of the network (in this case, the upper limit is 41 pixels). One can see that the proposed internal disparity loss points out where the error occurred.}
    \label{fig:apd:loss1}
\end{figure}

\begin{figure}
    \centering
    \includegraphics[width=\linewidth]{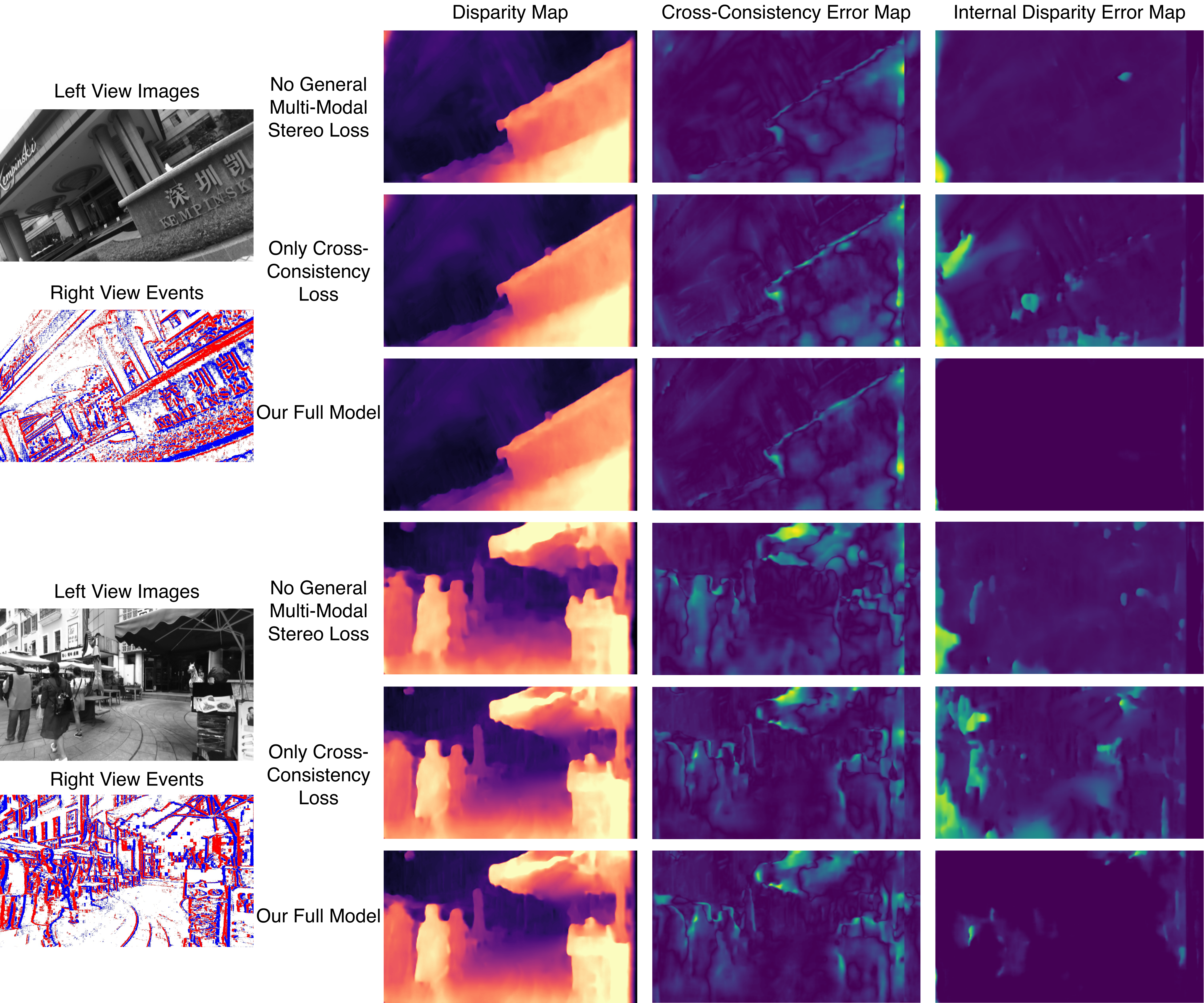}
    \caption{Visualization results of the proposed loss functions.}
    \label{fig:apd:loss2}
\end{figure}

We provide more results for the proposed general multi-modal stereo loss functions.
We first show the convergence curves with different loss functions and metrics in \figurename~\ref{fig:loss_curves}.
It can be seen that the proposed loss functions enable the model to improve its performance through self-supervised learning continuously.
The model without the proposed losses only provides a good guide initially, but when optimizing continuously, it does not match the purpose we want to achieve.
Since the gradient structure loss will be numerically unstable when the difference between the two images is large, some methods will fail halfway.
Although the internal disparity loss only provides simple, naive supervision, it has successfully improved the performance.
We will further understand these two loss functions through visualization results.
\figurename~\ref{fig:apd:loss1} and \figurename~\ref{fig:apd:loss2} show some error map visualization results.
One can first observe from \figurename~\ref{fig:apd:loss2} that the proposed cross-consistency loss helps outline the edge and shape of the disparity.
The proposed cross-consistency loss promotes the consistency of shapes between different views and provides additional information for training.
We can also see from the cross-consistency error maps and internal disparity error maps that the introduction of these losses reduces the degree of these inconsistencies, especially for the internal disparity loss.
\figurename~\ref{fig:apd:loss1} shows a failure case and also show how the internal disparity loss works.
In this case, the disparity of the front object exceeds the upper limit of the network (we set the max disparity to be 41 pixels).
The internal disparity loss reveals the failure area.

\section{Alternative Framework}

\begin{figure}
    \centering
    \includegraphics[width=\linewidth]{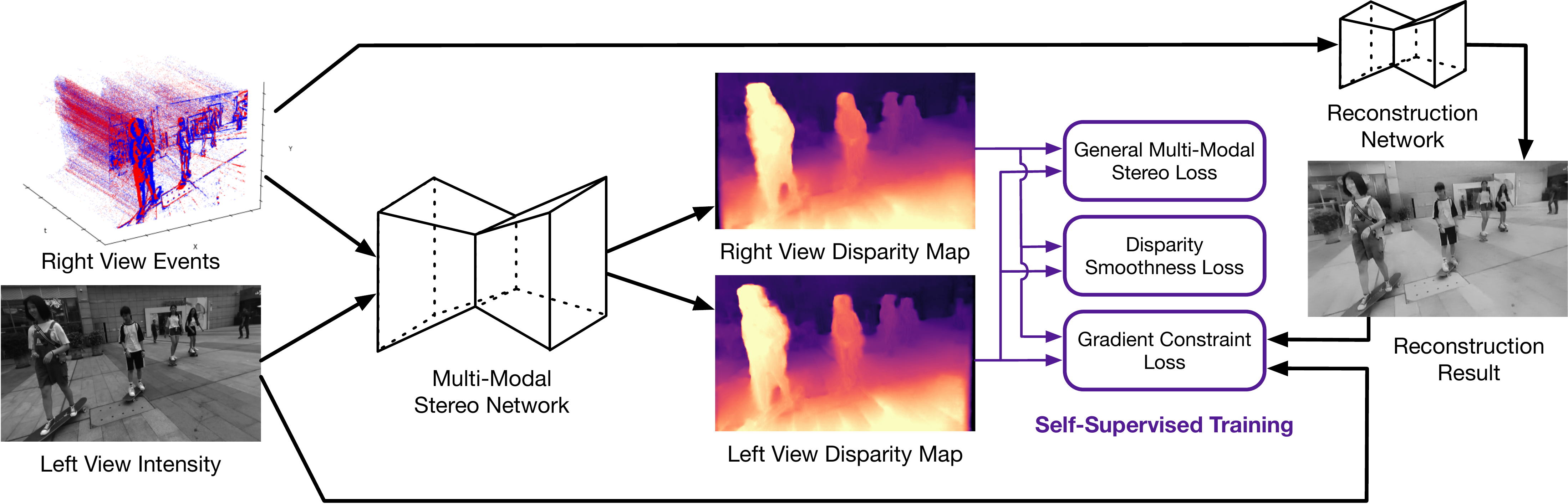}
    \caption{An alternative framework to the proposed self-supervised learning method. The stereo network takes the right event voxel and the left intensity image as input directly.}
    \label{fig:apd:framework}
\end{figure}

We also present an alternative framework where the stereo network takes the right event voxel and the left intensity image as input directly.
The framework is shown in \figurename~\ref{fig:apd:framework}.
In this framework, the network can also be the convolutional multi-modal stereo network.
However, using convolution to process event voxel directly tends to bring poor results.
We did not use this alternative since existing convolutional networks would be significantly better at processing images.
But this alternative shows that our general multi-modal stereo consistency loss can be generalized to a wider range of application scenarios.

%
%



\end{spacing}
\end{document}